\documentclass{article}

\usepackage{microtype}
\usepackage{graphicx}
\usepackage{subfigure}
\usepackage{booktabs} % for professional tables

\usepackage{hyperref}

\usepackage[accepted]{icml2024}

\usepackage{amsmath}
\usepackage{amssymb}
\usepackage{mathtools}
\usepackage{amsthm}

\usepackage[capitalize,noabbrev]{cleveref}

\theoremstyle{plain}

\theoremstyle{definition}

\theoremstyle{remark}

\usepackage[textsize=tiny]{todonotes}

\icmltitlerunning{video-SALMONN: Speech-Enhanced Audio-Visual Large Language Models}
\begin{document}

\twocolumn[
\icmltitle{video-SALMONN: Speech-Enhanced Audio-Visual Large Language Models}
%\icmltitle{video-SALMONN: Speech-Enabled Audio-Visual Large Language Models}
%\icmltitle{video-SALMONN: Audio-Visual Large Language Models with Enhanced Speech Abilities}
%\icmltitle{video-SALMONN: Audio-Visual Large Language Models Can Understand Not Only Sound and Music, but Also Speech}


\icmlsetsymbol{equal}{*}
%\icmlsetsymbol{xxx}{*}

\begin{icmlauthorlist}
\icmlauthor{Guangzhi Sun}{equal,yyy}
\icmlauthor{Wenyi Yu}{equal,yyy}
\icmlauthor{Changli Tang}{equal,yyy}
\icmlauthor{Xianzhao Chen}{comp}
\icmlauthor{Tian Tan}{comp}
\icmlauthor{Wei Li}{comp}
\icmlauthor{Lu Lu}{comp}
\icmlauthor{Zejun Ma}{comp}
\icmlauthor{Yuxuan Wang}{comp}
\icmlauthor{Chao Zhang}{yyy}
\end{icmlauthorlist}

\icmlaffiliation{yyy}{Department of Electronic Engineering, Tsinghua University}
\icmlaffiliation{comp}{ByteDance Ltd}

\icmlcorrespondingauthor{Chao Zhang}{cz277@tsinghua.edu.cn}

\icmlkeywords{Machine Learning, ICML}

\vskip 0.3in
]



\printAffiliationsAndNotice{\icmlEqualContribution} % otherwise use the standard text.

\begin{abstract}
%Understanding speech content as an element of video information in audio-visual large language models (av-LLM) is a crucial yet understudied aspect. 
Speech understanding as an element of the more generic video understanding using audio-visual large language models (av-LLMs) is a crucial yet understudied aspect. 
This paper proposes video-SALMONN, a single end-to-end av-LLM for video processing, which can understand not only visual frame sequences, audio events and music, but speech as well. To obtain fine-grained temporal information required by speech understanding, while keeping efficient for other video elements, this paper proposes a novel multi-resolution causal Q-Former (MRC Q-Former) structure to connect pre-trained audio-visual encoders and the backbone large language model. Moreover, dedicated training approaches including the diversity loss and the unpaired audio-visual mixed training scheme are proposed to avoid frames or modality dominance. On the introduced speech-audio-visual evaluation benchmark, video-SALMONN achieves more than 25\% absolute accuracy improvements on the video-QA task 
% due to the modelling of multi-resolution temporal information, 
and over 30\% absolute accuracy improvements on audio-visual QA tasks with human speech. In addition, video-SALMONN demonstrates remarkable video comprehension and reasoning abilities on tasks that are unprecedented by other av-LLMs. Our training code and model checkpoints are available at \texttt{\url{https://github.com/bytedance/SALMONN/}}. 
\end{abstract}

\section{Introduction}
\label{sec:intro}
Text-based large language models (LLMs) \citep{gpt3,llama,vicuna,palm2,glm} have demonstrated remarkable performance in many natural language processing tasks, especially achieving human-level capabilities in reasoning and comprehension \citep{gpt4}. Meanwhile, instruction fine-tuning \citep{flant5,Ouyang0JAWMZASR22,peng2023instruction}, where data is organised as paired user instructions (or prompts) and reference responses, has emerged as a training paradigm that enables LLMs to perform tasks by following open-ended natural language instructions from non-expert users. Recently, there has been a burgeoning research interest in equipping LLMs with visual and auditory perception abilities, resulting in a range of visual \citep{blip2,flamingo,instructblip,videochatgpt,videollm,lavila,lynx,valley}, audio \citep{gong2023listen,speechgpt,audiopalm,salmonn},
and audio-visual LLMs (av-LLMs) \citep{pandagpt,videollama,macaw,bubo,xllm,avllm,mirasol}.

Despite av-LLMs' prosperity, speech, as a primary carrier of human language in videos, is considerably under-explored in these models. Complementary to non-speech audio events and natural images, speech provides direct and abundant linguistic and semantic information, making it indispensable for comprehensive video understanding. Speech signals also include rich paralinguistic information, such as the tone and pitch of voice, which is often hard to textualise precisely but necessary to understand the underlying meanings and emotions.
Additionally, there exist diverse speaker attributes in speech, which are tedious and difficult to transcribe using separate systems but essential for video understanding (see Fig. \ref{fig:eg8}), including the speaker's age, gender, accent and identity \textit{etc}. To avoid building complex cascaded systems, it is desired to recognise and understand all of the aforementioned speech attributes in videos in a fully end-to-end and integrated way with av-LLMs. 
Nevertheless, enhancing general-purposed av-LLMs with speech is very challenging, which requires temporally fine-grained modelling while intricately interacting with other modalities at both coarse (\textit{e.g.} video topics) and fine (\textit{e.g.} lip movements) time scales. This necessitates the design of specialised fine-grained multi-resolution approaches to address this challenge.

To this end, we propose video-SALMONN (\textbf{s}peech \textbf{a}udio \textbf{l}anguage \textbf{m}usic \textbf{o}pen \textbf{n}eural \textbf{n}etwork), a speech-enhanced av-LLM for short video understanding. By resembling the audio encoder structure of the SALMONN~\cite{salmonn} LLM with generic hearing abilities and incorporating an additional visual encoder, video-SALMONN enables video inputs with natural image, visual frame sequence, speech, audio events, and music elements, covering all basic elements in general video data. 
%To our knowledge, video-SALMONN is the first multimodal LLM to achieve general video understanding abilities.
The core of video-SALMONN is a multi-resolution causal (MRC) Q-Former structure aligning time-synchronised audio-visual input features with text representation space at three different temporal scales, which meets the requirements of tasks relying on different video elements. 
To reinforce the temporal causal relations of events among successive video frames, a causal self-attention structure with a special causal mask is included in the MRC Q-Former. 
Further, to avoid the dominance of a specific frame or a single modality in the video, video-SALMONN is trained using a proposed diversity loss together with a new unpaired audio-visual mixing strategy.
% As a result, video-SALMONN is the first av-LLM tailored to achieve general video understanding to our knowledge. 
To our knowledge, video-SALMONN is the first av-LLM tailored to achieve general video understanding.

% Accurate multilingual video understanding in video-SALMONN is achieved via a Speech-enhanced Multilingual Audio-visual training (SMAT) scheme, where a vast amount of audio-visual English data is mixed with speech data from other languages during instruction tuning, which also alleviates the scarcity of annotated data in other languages. In the SMAT scheme, an unpaired audio and visual mixing strategy is adopted, increasing versatility and avoiding modality dominance. 
% Especially in multilingual activation tuning\cite{salmonn}, this mixed training naturally integrates the originally independent language-dependent speech and language-independent audio and video, thereby achieving both cross-lingual and cross-modal training.

To comprehensively evaluate the general video understanding abilities, we introduce the speech-audio-visual evaluation (SAVE) benchmark containing six open-source representative single-modal tasks and four open-source audio-visual tasks. video-SALMONN is the only av-LLM
% one of its type 
that can achieve tasks relying on speech elements, such as audio-visual speech recognition (AVSR) and speech-content-based QA. 
%such as automatic speech recognition (ASR), audio-visual speech recognition (AVSR) and speech-content-based QA.  
% With a particular focus on video with speech, MAVE contains two newly curated open-source video QA datasets for English and Chinese videos respectively. 
On the single-modal tasks, video-SALMONN achieves a remarkably 25\% accuracy improvement in Video QA, a question answering (QA) task 
focusing on temporal causal reasoning % based on visual frame sequence inputs, 
compared to a strong InstructBLIP baseline~\cite{instructblip}.
%, while considerably surpassing strong av-LLM baselines on non-English tasks. 
On audio-visual tasks, video-SALMONN has shown large performance improvements, \textit{e.g.} over 30\% absolute accuracy improvement on audio-visual QA dataset. 
%Moreover, we discover non-English speech understanding as an emergent ability for video-SALMONN trained on English-only data.
The main contributions are summarised as follows.
\vspace{-0.3cm}
\begin{itemize}
    \item We propose video-SALMONN, a speech-enhanced av-LLM. To our knowledge, video-SALMONN is the first single LLM-centric model that can handle video along with both speech and non-speech audio inputs.
    \vspace{-0.2cm}
    \item We propose the MRC Q-Former structure as a multi-resolution modality aligner for video-SALMONN, which lays a solid foundation for the joint speech-audio-visual information extraction in videos. 
    \vspace{-0.2cm}
    \item We propose the diversity loss and mixed training scheme to achieve a better balance of features from different frames and modalities. 
    \vspace{-0.2cm}
    %avoid features from a certain frame or modality dominating the multimodal temporal sequence processing in video understanding. 
     %avoid frame and modality dominance in video understanding.
    \item video-SALMONN achieves superior performance on the SAVE benchmark, especially in audio-visual tasks requiring speech understanding and causal reasoning.
\end{itemize}

\begin{figure*}[t]
    \centering
    \includegraphics[scale=0.23]{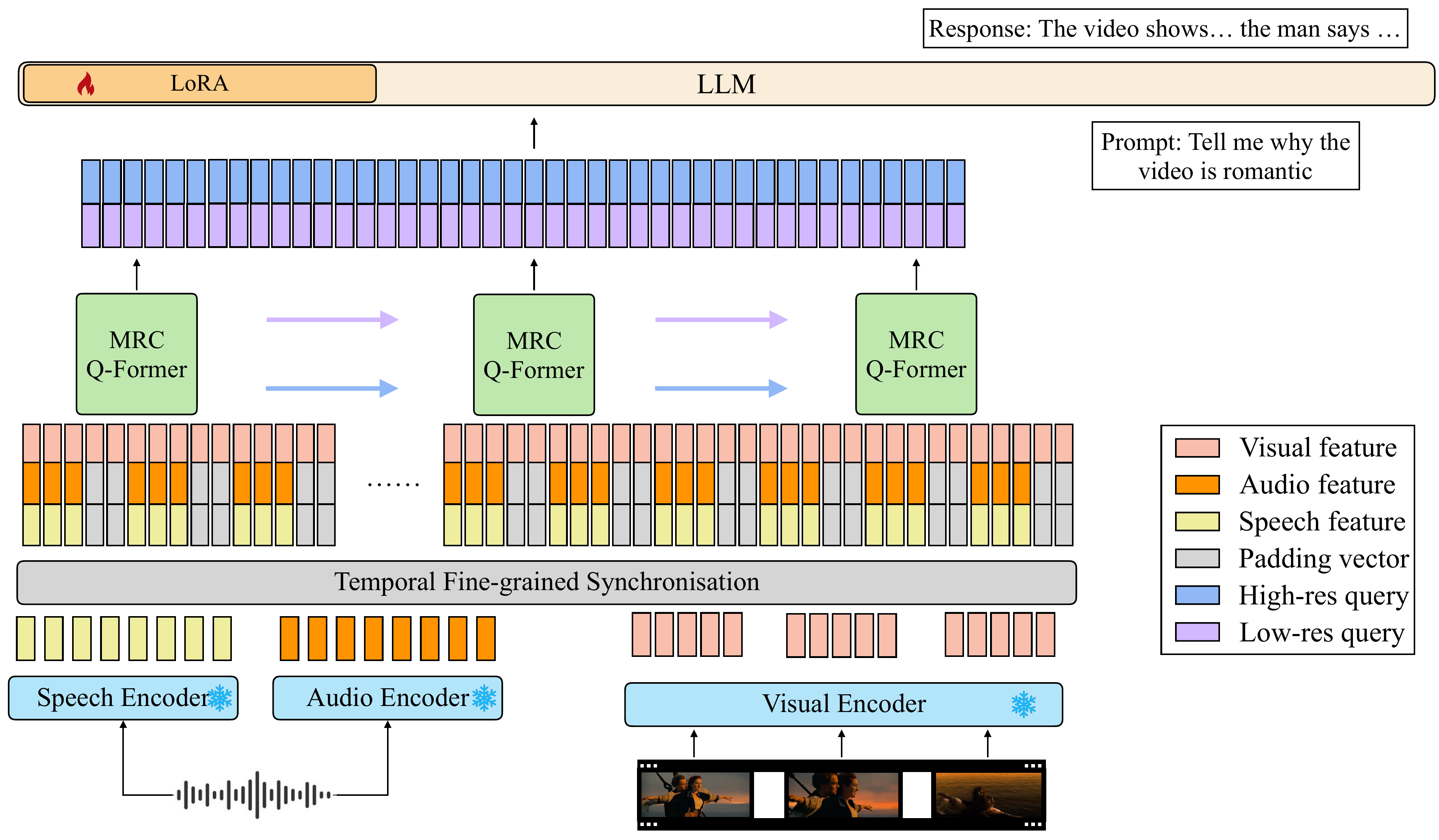}
    \vspace{-0.3cm}
    \caption{The model structure of video-SALMONN using fine-grained audio-visual joint representations. Audio and visual input streams are encoded into sequences of features with individual encoders that are not updated during training, and the features are temporally synchronised and processed by the proposed multi-resolution causal (MRC) Q-Former operating at different time scales.}
    \vspace{-0.3cm}
    \label{fig:avsalmonn}
\end{figure*}
\section{Related Work}

The work most closely related to video-SALMONN is Video-LLaMA \citep{videollama}, Macaw-LLM \citep{macaw}, X-LLM \citep{xllm} and also work proposed by \citet{audiovisual,vast}, as all of them used LLMs for cross-modal understanding based on general non-silent video inputs (referred to as audio-visual sequence in this paper).
X-LLM supports video with Chinese speech inputs, but doesn't support audio events and music. 
Video-LLaMA employs an additional video Q-Former to encode features of several equally-spaced frames extracted using a BLIP2 \citep{blip2} image encoder. Macaw-LLM adopted a similar approach and used three separate encoders for image, video and non-speech audio events. 
Both Video-LLaMA and Macaw-LLM consider only non-speech audio events, and the audio encoders in the two models are the ImageBind \citep{imagebind} and Whisper \citep{whisper} model encoders respectively. 
While both methods involve the fusion of audio and visual feature streams, the two streams are sparsely pooled and processed rather independently, which removes fine-grained audio-visual interactions at each time step.
Compared to Video-LLaMA and Macaw-LLM, video-SALMONN understands speech in a video and reserves fine-grained modality interactions that are common in general non-silent videos. 
This leads to an emphasis on causal modality synchronisation across time and allows more content-based cross-modal interactions.

As an alternative to include speech modelling in av-LLM, speech content can be extracted using an external automatic speech recognition (ASR) system and fed into the av-LLM as textual subtitle inputs \cite{vast}. 
However, this approach ignores the rich paralinguistic and speaker information embedded in speech, unless they are also extracted using external systems. 
Rich transcription (RT) is a long-standing research problem targeting extracting abundant information from speech signals \cite{rt04,rt05,rt06,rt07} that used to be tackled as several separate tasks, such as ASR, speaker diarisation and emotion recognition \textit{etc}. 
In contrast, video-SALMONN unifies those hearing ability tasks together with visual perception abilities using a single end-to-end model.

%To avoid ignoring the rich paralinguistic and speaker information embedded in speech, each 
%Rich transcription (RT) is a long-standing research problem targeting extracting abundant information from speech signals \cite{rt04,rt05,rt06,rt07} that used to be tackled using a cascaded pipeline with several separate systems, such as ASR, speaker diarisation and emotion recognition systems \textit{etc}. 

%Rich transcription (RT) extracting abundant information (content, speaker, paralinguistics, \textit{etc.}) from speech signal has been a long-standing research problem \cite{rt04,rt05,rt06,rt07} that used to be tackled with separate systems for each task. In contrast, video-SALMONN unifies those tasks together with visual perception abilities in a completely end-to-end model.

Our work is based on the Q-Former structure to fuse the audio and visual modalities and to align with the text representation space \citep{blip2,instructblip}. While Q-Former has been primarily proposed for visual information extraction, it also performs remarkably in extracting auditory features for generic audio understanding in SALMONN~\citep{yuwenyi,tangchangli,salmonn}.
In addition, various types of modality aligners have been studied, such as the cross-attention mechanism \citep{flamingo}, pre-trained multimodal embeddings, \citep{imagebind} and temporal and spatial pooling \citep{videochatgpt} \textit{etc}. Different from these approaches, our proposed MRC Q-Former used in video-SALMONN pays particular attention to the sequential nature of video and the multi-resolution information of the input feature streams suitable for the understanding of different video elements. This work is a revision of an unpublished work of ours \cite{favor}, which is the first study to explore video understanding with general audio (audio event, speech and music \textit{etc.}).

%with the model structure and training methods dedicated to audio-visual understanding.

%Our work is based on the Q-Former structure to fuse the audio and visual modalities and to align with the text representation space \citep{blip2,instructblip}. While Q-Former has been primarily proposed for visual information extraction, it also performs remarkably in extracting auditory features for generic audio understanding in SALMONN\citep{yuwenyi,tangchangli,salmonn}. In addition to Q-Former, various types of modality aligners have been studied, such as the cross-attention mechanism \citep{flamingo}, pre-trained multimodal embeddings, \citep{imagebind} and temporal and spatial pooling \citep{videochatgpt}. Different from standard Q-Former approaches, our proposed MRC Q-Former used in video-SALMONN pays particular attention to the sequential nature and the multi-resolution information of the input feature streams with the model structure and training methods dedicated to audio-visual understanding.

\section{video-SALMONN}
\label{sec:modelstructure}
This section introduces the structure and the training approach for video-SALMONN. As shown in Fig. \ref{fig:avsalmonn}, key components include the synchronisation module and the MRC Q-Former. First, visual (image or video), speech and non-speech audio are encoded using corresponding pre-trained encoders. The visual encoder converts the input image into a certain number of vectors via the image encoder from InstructBLIP \citep{blip2}. When video input is given, the visual encoder encodes each video frame separately as a sequence of images at a 2 Hz frame rate, and the output image features are concatenated along the temporal dimension to form a sequence of visual frames. Following SALMONN \cite{salmonn}, Whisper \cite{whisper} encoder and BEATs \cite{beats} encoder are adopted to encode speech and non-speech audio respectively from the same audio stream at 50 Hz spectrogram frame rate.

\subsection{Temporal Fine-grained Synchronisation}
%When both audio and visual inputs are present, the two encoded feature sequences are sent to the temporal synchronisation module to obtain the time-synchronised feature sequences. Since video is sampled at a lower frame rate than audio, the audio and visual frames are synchronised at each video frame (\textit{i.e.} every 0.5 seconds), with zero padding to make both sequences have equal lengths.
%Note that higher frequencies of visual frames are also supported which requires higher computation and storage costs. 
%The synchronised frame-level encoder outputs $\mathbf{h}^{\text{S}}_t$, $\mathbf{h}^{\text{A}}_t$ and $\mathbf{h}^{\text{V}}_t$ for speech, audio and video respectively are then concatenated along the feature dimension to obtain the combined representation $\mathbf{h}^{\text{SAV}}_t$. That is,
%that are further encoded and compressed by the SW Q-Former.
When both audio and visual inputs are present, the encoded feature sequences are sent to the temporal synchronisation module to obtain the time-synchronised feature sequences. Since video is sampled at a lower frame rate than audio, the audio and visual frames are synchronised at each video frame (\textit{i.e.} every 0.5 seconds), with zero padding to make both sequences have equal lengths.
Note that higher frequencies of visual frames are also supported which requires higher computation and storage costs. 
$\mathbf{h}^{\text{S}}_t$, $\mathbf{h}^{\text{A}}_t$ and $\mathbf{h}^{\text{V}}_t$, the synchronised frame-level outputs at step $t$ from the Whisper speech encoder, BEATs audio encoder and InstructBLIP video encoder, are concatenated along the feature dimension to obtain the combined representation $\mathbf{h}^{\text{SAV}}_t$. That is,
\begin{equation}
    \mathbf{h}^{\text{SAV}}_t = \text{Concat}(\mathbf{s}_{t}, \mathbf{a}_t, \mathbf{v}_t),
    \label{eq:sync}
\end{equation}
%where Concat$(\cdot)$ represents the concatenation along the feature dimension. Note that in cases when only one input modality is present, the other modality is filled with a sequence of zero padding of the same sequence length. 
%While an image alone is treated as a single frame, when paired audio input exists, such as images with spoken captions \citep{spokencoco}, each image is duplicated as if it were a video input with a matched length to the audio input.
where Concat$(\cdot)$ represents the concatenation along the feature dimension and $\mathbf{W}$ is a projection weight matrix. Note that in cases when audio input is missing, $\mathbf{s}_t$ and $\mathbf{a}_t$ are replaced with a sequence of zero padding of the same sequence length, and \textit{vice versa}. 
While an image alone is treated as a single frame, when paired audio input exists, such as images with spoken captions \citep{spokencoco}, each image is duplicated as if it were a video input with a matched length to the audio input.

\subsection{MRC Q-Former}
\begin{figure}[t]
    \centering
    \includegraphics[scale=0.19]{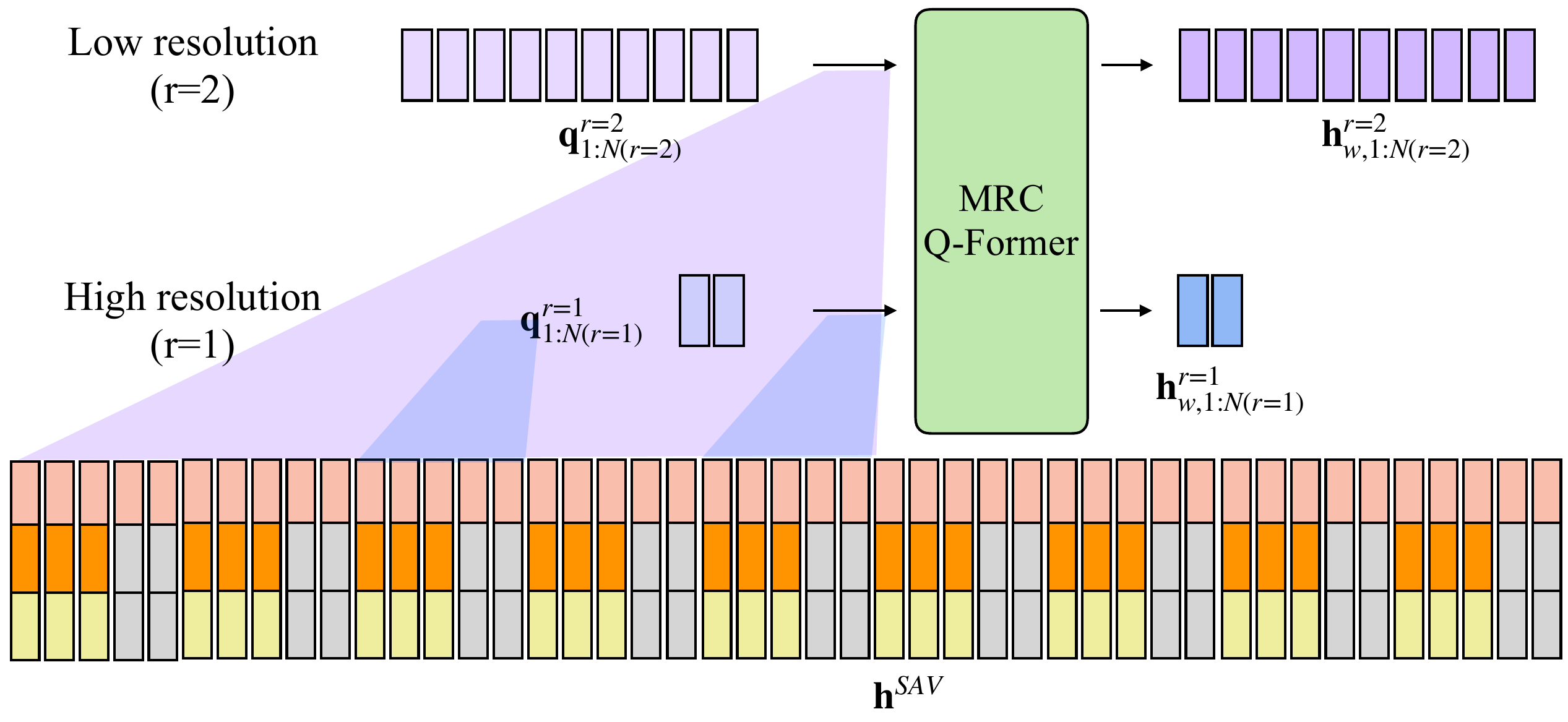}
    \vspace{-0.6cm}
    \caption{Illustration of the MRC Q-Former structure with two levels of temporal resolutions. The high-resolution sliding window covers $k=5$ input features with two query vectors and the low-resolution Q-Former covers $k=25$ with 10 query vectors.}
    \vspace{-0.3cm}
    \label{fig:mrcqformer}
\end{figure}

The MRC Q-Former extracts audio-visual features from variable-length inputs at different temporal resolutions. The detailed structure is shown in Fig. \ref{fig:mrcqformer}. First, the synchronised input stream is divided into fixed-length windows at multiple different resolutions, \textit{e.g.} spanning every 1, 5 or 10 seconds. Then, at each resolution level $r$, based on $N(r)$ trainable input query vectors, the MRC Q-Former is applied to convert features in each sliding window into $N(r)\in \mathbb{N}^+$ output query vectors carrying the audio-visual joint information. That is,
% as shown in Eqn. (\ref{eq:swqformer}).
\begin{equation}
    \mathbf{h}^{(r)}_{w, 1:N(r)} = \text{Q-Former}_\text{MRC}(\mathbf{h}^{\text{SAV}}_{t:t+k(r)};\mathbf{q}^{(r)}_{1:N(r)}),
    \label{eq:swqformer}
\end{equation}
%where $w$ is the window index and $k(r)$ is the number of video frames in each window at resolution level $r$, and $\text{Q-Former}_\text{MRC}(\cdot)$ denotes the Q-Former computation \cite{blip2}. The output query vectors are $\mathbf{h}^{r}_{w, 1:N(r)}$. If the input sequence length of the MRC Q-Former is $T$, the number of sliding windows $W({r})\in \mathbb{N}^+$ becomes $\lceil T/k(r)\rceil$, and the overall output sequence length from the MRC Q-Former will be $W(r)\times N(r)$. The sliding window design enables the length of the input sequence to vary according to the input feature sequence lengths. It hence achieves a better balance between the degree of information reserved and the computation and storage costs than using a single Q-Former for the entire sequence.
where $w$ is the window index and $k(r)$ is the number of input video frames in each window at resolution level $r$, and $\text{Q-Former}_\text{MRC}(\cdot)$ denotes the Q-Former computation \cite{blip2}. The output query vectors are $\mathbf{h}^{(r)}_{w, 1:N(r)}$. If the input sequence length of the MRC Q-Former is $T$, the number of sliding windows $W({r})\in \mathbb{N}^+$ becomes $\lceil T/k(r)\rceil$, and the overall output sequence length from the MRC Q-Former will be $W(r)\times N(r)$. The sliding window design enables the length of the input sequence to vary according to the input feature sequence lengths. It hence achieves a better balance between the degree of information reserved and the computation and storage costs than using a single Q-Former for the entire sequence.

%This operation is repeated for all resolution levels with resolution-specific query vectors. We ensure that Q-Former output at different resolutions can be synchronised by enforcing Eqn. (\ref{eq:constrain}) where $C$ is a hyper-parameter representing the total number of output query vectors sent to the LLM.
This operation is repeated for all resolution levels with the resolution-specific query vectors. We ensure that Q-Former output at different resolutions can be synchronised by enforcing Eqn. (\ref{eq:constrain}), where $C$ is a hyper-parameter representing the total number of output query vectors sent to the LLM:
\begin{equation}
    W({r})\times N(r) = C.
    \label{eq:constrain}
\end{equation}
%When applying smaller windows for finer time scales, a smaller number of query vectors is used for a reduced total information capacity, and \textit{vice versa}. Note that while query vectors are different at different resolutions, we share the parameters for the MRC Q-Former across all resolution levels as the task of modality alignment is the same.
When applying smaller windows for finer time scales, a smaller number of query vectors is used for a reduced information capacity, and \textit{vice versa}. Note that while keeping the query vectors different for different resolutions, the rest of the MRC Q-Former parameters are shared across all resolution levels as the task of modality alignment is the same. Output query vectors at all resolution levels are combined using a projection layer before sending them to the LLM.
\begin{equation}
\mathbf{H}=\mathbf{W}^{(1)}\mathbf{H}^{(1)}+\dots+\mathbf{W}^{(R)}\mathbf{H}^{(R)}
    \label{eq:generate}
\end{equation}
where each $\mathbf{H}^{(r)}=[\mathbf{h}^{(r)}_{w, 1:N(r)}]^{\lceil T/k(r)\rceil}_{w=1}\in \mathbb{R}^{C\times D}$ includes output query vectors at resolution level $r$, $D$ is the dimension of output query vectors, and $\mathbf{W}^{(r)}\in\mathbb{R}^{D\times E}$ projects output query vectors to the LLM input embedding dimension $E$.
Finally, the LLM backbone generates output based on the projected query vectors $\mathbf{H}$ and the content of the prompt $\mathbf{c}_1,\mathbf{c}_2,\ldots,\mathbf{c}_M$ by
\begin{equation}
\vspace{-0.3cm}
    \hat{\mathbf{Y}}=\operatorname*{argmax}_\textbf{Y} P(\mathbf{Y}|\mathbf{H},\mathbf{c}_{1:M}).
    \label{eq:generate}
\end{equation}
% concatenated output query vectors of all windows projected to the LLM input dimension to form the input to the LLM. 

% where $\mathbf{W^{(r)}}$ projects output query vectors to the LLM input dimension The response sequence $\hat{\mathbf{Y}}$ can be generated as follows: 
% \begin{equation}
%     \hat{\mathbf{Y}}=\operatorname*{argmax}_\textbf{Y} P(\mathbf{Y}|\text{Concat}(\mathbf{H}^{(1)},\dots,\mathbf{H}^{(R)}),\mathbf{c}_{1:M}),
%     \label{eq:generate}
% \end{equation}
%\begin{equation}
%    \hat{\mathbf{Y}}=\operatorname*{argmax}_\textbf{Y} P(\mathbf{Y}|\mathbf{H},\mathbf{c}_{1:M}),
%    \label{eq:generate}
%\end{equation}
% where $\mathbf{c}_1,\mathbf{c}_2,\ldots,\mathbf{c}_M$ are the contents of the prompt. Each $\mathbf{H}^{(r)}=[\mathbf{h}^{(r)}_{w, 1:N(r)}]^{\lceil T/k(r)\rceil}_{w=1}\in \mathbb{R}^{C\times D}$ includes output query vectors at resolution level $r$ where $D$ is the dimension of output query vectors.
%, and Concat$(\cdot)$ denotes the concatenation along the feature dimension. 
%The system is trained in an end-to-end fashion using the cross-entropy loss on the LLM response.
% The system is trained in an end-to-end fashion.

\subsubsection{Causal Structure}
\begin{figure}[t]
    \centering
    \includegraphics[scale=0.23]{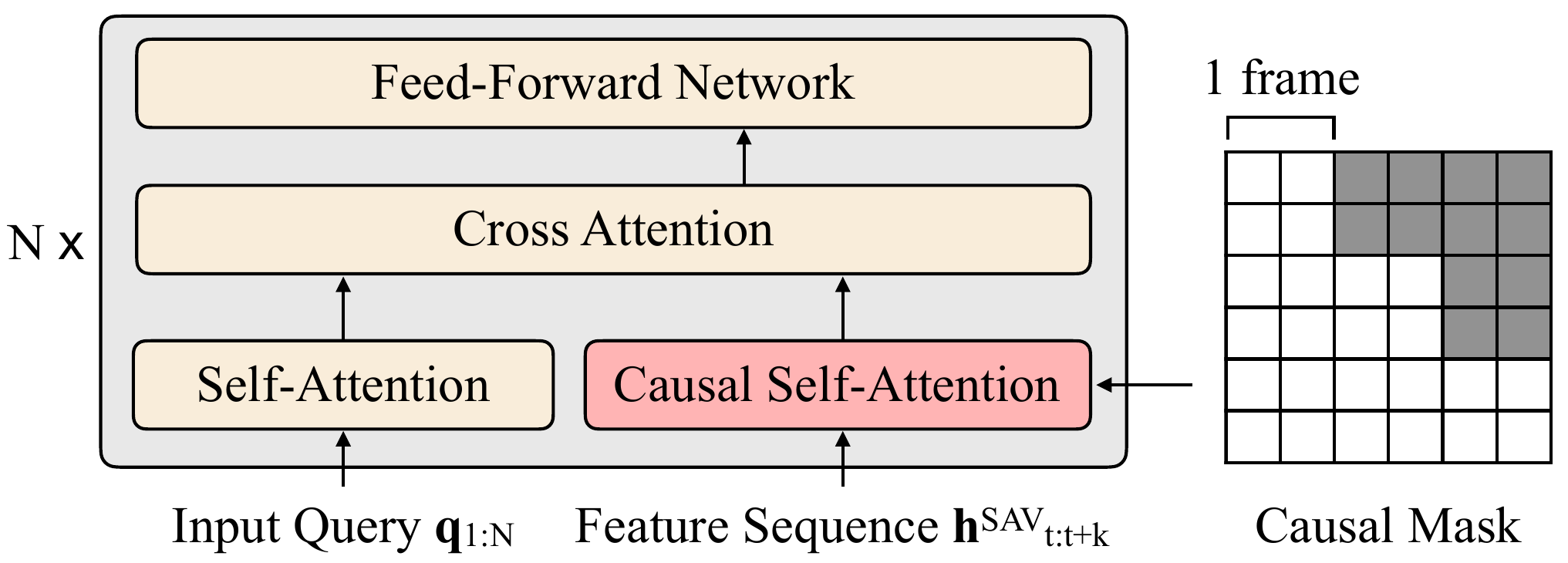}
    \vspace{-0.2cm}
    \caption{The causal attention module in the MRC Q-Former with a block-wise triangular causal mask (grey cells are masked). The number of features per frame here is two as an example.}
    % \vspace{-0.3cm}
    \label{fig:swqformer}
\end{figure}

The proposed MRC Q-Former adopts a causal structure as shown in Fig. \ref{fig:swqformer}. To capture the causal temporal correlation among frames that are extracted independently, an additional causal self-attention module is added to the standard Q-Former structure, indicated by the red block in Fig. \ref{fig:swqformer}.

With the causal attention module, the encoding of one specific frame also includes the information of all previous frames carried in an auto-regressive way.
This is particularly beneficial for causal reasoning questions, such as the ``what happens next'' questions \citep{nextqa}. Such questions are sometimes difficult to learn using only the positional embeddings.

% \subsection{Training Objectives}

\subsection{System Training}
The training data of video tasks such as video QA usually only requires one or two keyframes, and the output queries tend to repeatedly capture the same information. Therefore, a novel diversity loss is proposed to encourage the MRC Q-Former to extract more diverse aspects of the input sequence. Specifically, the diversity loss is formulated as: 
\begin{equation}
    \mathcal{L}_\text{diverse} = \sum_{r=2}^{R}\sum_{w=1}^{W(r)} \sum_{i=1}^{N} \sum_{j=1 {, j\neq i}}^{N} \text{sim}(\mathbf{h}_{w,i}^{(r)}, \mathbf{h}_{w,j}^{(r)})
    \label{eq:div}
\end{equation}
where $W(r)$ and $N(r)$ are the total number of windows and the number of output queries of each window at resolution level $r$ respectively, and sim$(\cdot)$ is the cosine similarity between two vectors. 
Cosine similarity is adopted since it is widely used for semantic similarity measurements, and in video-SALMONN, the output queries are aligned with a semantic space of the LLM input token representations. 
This choice is also supported by the fact that the modulus of the output query tokens is very similar due to the layer normalisation operation of the MRC Q-Former. Note that the diversity loss is only needed at the low-resolution levels where there are enough frames in a window to extract diverse information.

% By encouraging different queries to focus on different aspects or frames of the input stream, the diversity loss forces the output query representations to be more spread in the text representation space. 

Overall, video-SALMONN is trained end-to-end using the cross-entropy (CE) loss and the diversity loss as shown below, where $\lambda$ controls the importance of the diversity loss. 
\begin{equation}
    \mathcal{L} = \mathcal{L}_\text{CE} + \lambda \mathcal{L}_\text{diverse},
    \label{eq:totalloss}
\end{equation}
Furthermore, to avoid modality dominance in the video, in addition to the small amount of paired audio-visual data, we propose the mixed training scheme where a portion of the training set is augmented with unpaired audio-visual data and the prompt combines the original tasks for audio and video. This way, the model is enforced to extract information from both audio and video inputs without relying on a dominant modality. This strategy improved the balance between different modalities and is a crucial factor leading to audio-visual understanding and co-reasoning abilities. 
% demonstrated in Fig. \ref{fig:eg1} to Fig. \ref{fig:eg6} in Appendix \ref{casestudy}.

\section{Experimental Setup}
\subsection{Speech-Audio-Visual Evaluation Benchmark}
\label{sec:mave}
\begin{table*}[t]
    \footnotesize
    \centering
        \caption{SAVE benchmark details, including the number of samples used for evaluation and metrics reported. Since TextVQA, GQA, NExT-QA and VGGSS test sets are large, randomly sampled subsets with enough samples for statistical significance were used for efficient evaluation. Zero-shot refers to both instruction and audio-visual inputs that are unseen in the training set. Note that Presentation-QA 
        % and Bilibili-QA are 
        is newly proposed AVQA test sets focusing on speech-audio-visual joint information.}
    \vspace{0.3cm}
    \begin{tabular}{llllll}
    \toprule
    \textbf{Task}     & \textbf{Test set} & \textbf{\#samples} & \textbf{Metrics} & {\textbf{Zero-shot}}\\
    \midrule
    ASR     & LibriSpeech test-clean \citep{librispeech} & 2620 & WER & No\\
    % & Commonvoice & - & Zh, De & WER & No\\
    AAC & AudioCaps test \citep{audiocaps} & 938 & SPIDEr & No \\
    IC & Flickr30k test \citep{flickr30k} & 1000 & CIDEr & Yes \\
    OCR & TextVQA test \citep{textvqa} & 1000 & Accuracy & Yes \\
    VQA & GQA test dev balanced \citep{gqa} & 1000 & Accuracy & Yes \\
    Video QA & NExT-QA test \citep{nextqa} & 1000 & Accuracy & Yes \\
    \midrule
    AVSR & How2 dev5 \citep{how2} & 500 & WER & No \\
    % & AVASR & - & Zh & WER & No \\
    % AVSD & AVSD val \citep{avsd} & 2000 & En & Accuracy & No \\
    AVQA & Ego4D \cite{ego4d} + Presentation-QA & 2000 & Accuracy & Yes \\
    % ISQA & TextVQA + GQA & 2000 & Accuracy & Yes \\
    AVSSD & VGGSS \citep{vggss,bubo} & 850 & Accuracy & Yes \\
    AVM & SpokenCOCO \citep{spokencoco} + VGGSS & 1000 & Accuracy & Yes \\
    \bottomrule
    \end{tabular}
    \label{tab:task}
\end{table*}
We introduce the SAVE benchmark to evaluate the performance of video-SALMONN. SAVE benchmark contains selected representative tasks for both single and multi-modal tasks. The six single-modal tasks included are ASR, automatic audio captioning (AAC), image captioning (IC), optical character recognition (OCR), visual question answer (VQA), and video question answer (Video QA), and the four audio-visual tasks spanning 6 datasets are audio-visual speech recognition (AVSR), audio-visual QA (AVQA), audio-visual matching (AVM) and audio-visual sound source detection (AVSSD).
% \added{In addition, we incorporate two widely used audio-visual benchmarks, the fine-grained audible video description (FAVD) \citep{favd} and the Vision-Audio-Language Omni-peRception (VALOR) \citep{valor,vast} benchmarks in our evaluation.}

In particular, we curate Ego4D-QA and Presentation-QA test sets to evaluate accuracy in audio-visual understanding with speech. The questions for the two sets are generated by prompting GPT-4 with video descriptions and ASR transcriptions for each video clip. Detailed examples for AVQA datasets are in Appendix \ref{bilibiliqa}. The SAVE benchmark is summarised in Table \ref{tab:task}, and details about evaluation metrics can be found in Appendix \ref{evaluation}. 

% While all other tasks already exist with open-source test sets, this paper proposes ISQA and AVM where audio-visual interaction is necessary. ISQA is the task where the question is in the audio and the answer can be found in the image. This test set is derived from the data used for OCR and VQA, where the questions are synthesised using a commercial text-to-speech synthesis system with a diverse range of speakers and styles. The text prompt is always ``answer the question in the audio about the image", while the LLM is required to first understand the question in the speech, and then answer it by looking at the image. 
This paper further proposes the AVM task where audio-visual interaction is necessary. AVM is the task of determining whether the given spoken description in the SpokenCOCO dataset \citep{spokencoco} matches the image, or whether the given audio clip is compatible with the given video chosen from the VGGSS dataset \citep{vggss}. AVSSD is another task that requires a strong binding of audio and visual modalities, as a single modality usually only provides partial information about the sound. 

\subsection{Model Configurations}

To validate video-SALMONN on the SAVE benchmark, the Vicuna-v1.5 \citep{vicuna} models (including 7B and 13B models, and 13B is the default option if not specified) is used as the LLM, Whisper \citep{whisper} large-v2 encoder as the speech encoder, BEATs \cite{beats} encoder as the audio encoder and InstructBLIP \citep{instructblip} vision Transformer (ViT) plus Q-Former as the visual encoder. The visual encoder outputs 32 feature vectors for each video frame (every 0.5 seconds), and the audio encoder outputs 50 feature vectors per second. 

The MRC Q-Former has two Transformer blocks with $D$=768-dim hidden states. By default, we adopt two different levels of resolution at 0.5-second and 5-second respectively, with the number of output query vectors being 3 and 30 for each window. The output query vectors of the MRC Q-Former are projected to $E$=5120-dim before being sent to the LLM. 
The LLM is adapted using the low-rank adaptation (LoRA) \citep{lora} method with rank 32. LoRA parameters of the attention query, key and value projections and feed-forward network weights are updated, which comprises 0.4\% of the total number of LLM parameters.

Whisper and InstructBLIP are used as the single-modality baseline systems for comparison. As video-SALMONN uses video data with different styles and focuses, to eliminate the discrepancy in training data and achieve fair comparisons, InstructBLIP is further fine-tuned on the same image and video training data as video-SALMONN. For each video clip, five equally-spaced frames were used resulting in 160 output queries. This is the same as the number of output queries used for 25-second videos in video-SALMONN. Video-LLaMA \citep{videollama} was used as the multimodal baseline where only the Vicuna-7B checkpoint was released for audio-visual input.
% \footnote{\url{https://github.com/DAMO-NLP-SG/Video-LLaMA.git}.}. 

\begin{table*}[t]
    \vspace{-0.3cm}
    \footnotesize
    \centering
    \caption{The SAVE benchmark single-modal task results. 
    If specified, InstructBLIP is fine-tuned on the training data of video-SALMONN (``InstructBLIP fine-tuned''). Evaluation metrics can be found in Appendix \ref{evaluation}. When using visual-only inputs, the other modality is masked during training and inference. Tasks unable to be performed are marked with ``-".}
    \vspace{0.2cm}
    \begin{tabular}{lcccccc}
    \toprule
     \textbf{Systems}    &  \textbf{ASR} $\downarrow$  & \textbf{AC} $\uparrow$ & \textbf{Video QA} $\uparrow$  & \textbf{IC} $\uparrow$ & \textbf{OCR} $\uparrow$ & \textbf{VQA} $\uparrow$ \\
    \midrule
    Whisper large-v2    & {2.9}\% & - & - & - & - & - \\
    InstructBLIP 13B \cite{instructblip} & - & - & 21.0\%  & 84.5 & 36.5\% & \textbf{48.9}\% \\
    InstructBLIP 13B fine-tuned & - & - & 24.7\% & 78.9 & 36.7\% & 45.6\% \\
    Video-LLaMA 7B \cite{videollama} & 100\%+ & 3.5 & 22.5\% & 22.0& 16.4\% & 15.1\%\\
    \midrule
    % video-SALMONN 13B (ours, audio-only) & \textbf{2.7}\% & - & - & - & - & - \\
    video-SALMONN 13B (ours, visual-only) & - & - & 44.8\%  & 74.0 & 34.2\% & 45.6\%\\
    video-SALMONN 7B (ours) & 4.1\% & 39.1 & 42.5\% & 78.1 & 34.6\% & 45.3\%  \\
    video-SALMONN 13B (ours) & \textbf{2.6}\% & \textbf{49.7} & \textbf{49.6}\% & \textbf{89.6} & \textbf{37.8}\% & {44.8}\% \\
    % video-SALMONN 13B + BEATs (ours) & 2.8\% & 46.2 & 46.6\% & 25.7 / 84.7 & 37.2\% & 45.1\%\\
    % \midrule
    % video-SALMONN-EN 13B v1.5 (ours) & \textbf{2.7}\% & 41.3 & 45.2\% & \textbf{90.2} / 26.6 & \textbf{38.3}\% & 45.3\% \\
    % video-SALMONN-EN 13B v1.5 + BEATs (ours) & \\
    \bottomrule
    \end{tabular}
    \label{tab:singlemod}
\end{table*}

\begin{table*}[t]
    \centering
    \footnotesize
    \caption{The SAVE benchmark audio-visual task results. If specified, InstructBLIP is fine-tuned on the training data of video-SALMONN (``InstructBLIP$\dag$''). The other modality is masked in both training and testing when using visual-only inputs. Tasks unable to be performed are marked with ``-''. We split AVQA into Ego4D-QA (E) and Presentation-QA (P).}
    \vspace{0.2cm}
    \begin{tabular}{lcccccc}
    \toprule
     \textbf{Systems}    &  \textbf{AVSR} $\downarrow$ $\uparrow$ & \textbf{AVQA (E)} $\uparrow$ & \textbf{AVQA (P)} $\uparrow$ & \textbf{AVSSD} $\uparrow$ & \textbf{AVM} $\uparrow$ \\
    \midrule
    Whisper large-v2    & 8.3\% & - & -  & - & - \\
    InstructBLIP 13B \cite{instructblip} & - & - & - & 1.1\% & - \\
    InstructBLIP$\dag$ 13B & - & - & - & 20.3\% & - \\
    Video-LLaMA 7B \cite{videollama} & - & 18.2\% & 21.3\% & 41.9\% & 52.3\% \\
    \midrule
    % video-SALMONN 13B (ours, audio-only) & 8.3\% & - & - & - & 34.7\% & - \\
    video-SALMONN 13B (ours, visual-only) & - & 35.0\% & 46.5\% & 23.5\% & - \\
    video-SALMONN 7B (ours) & 8.7\% & 36.2\% & 41.3\% & \textbf{50.5}\% & 74.3\% \\
    video-SALMONN 13B (ours) & \textbf{7.7}\% & \textbf{49.8}\% & \textbf{70.5}\% & {47.6}\% & \textbf{79.7}\% \\
    % video-SALMONN 13B + BEATs (ours) & 8.2\% & 51.4\% \\
    % \midrule
    % video-SALMONN-EN 13B v1.5 (ours) & \textbf{8.0}\% & & & 25.5\% & \\
    % video-SALMONN-EN 13B v1.5 + BEATs (ours) & \\
    \bottomrule
    \end{tabular}
    \vspace{-0.2cm}
    \label{tab:audiovisual}
\end{table*}

\subsection{Training Data and Specifications}

Multi-task instruction fine-tuning is used to train model parameters of MRC Q-Former and LoRA in video-SALMONN. Training data contains both single-modal and audio-visual paired data. For audio-only tasks, LibriSpeech train-clean-100 and train-clean-360 sets are used for ASR, and AudioCaps are used for AAC. For visual-only tasks. A mixture of LLAVA-150k \citep{llava} image QA data, OCRVQA OCR data \citep{ocrvqa}, TextCaps \cite{textcaps} image caption data, NExT-QA\footnote{The instruction format (\textit{i.e.} multiple choice questions) and videos for testing are all unseen for NExT-QA, hence zero-shot.} video QA training data \citep{nextqa}, 5000 samples from COCO train2014 data with spoken captions \citep{coco} as well as 11k samples from VideoChat \citep{videochat} are used. For audio-visual tasks, randomly selected 600-hour Ego4D video captioning data \citep{ego4d}, How2 300-hour training set AVSR data and audio-visual scene-aware dialogue (AVSD) training set are used. The entire training data only contains 1M samples with fewer than 300k video samples, with only publicly available datasets. Details about the training data can be found in Appendix \ref{testset}.

\begin{table*}[t]
    \centering
    \footnotesize
    \caption{Ablation studies on the core components of video-SALMONN based on single modal and audio-visual tasks. Each row represents removing one or more components with other parts remaining the same. Note the last row is equivalent to Video-LLaMA with the same training data, high frame rate video, speech encoder and LoRA, and the comparison to complete video-SALMONN directly reflected the benefit of the proposed structural and training design. AVQA takes the average among the two datasets.}
    \vspace{0.2cm}
    \begin{tabular}{lcccccc}
    \toprule
    \textbf{Systems}     & \textbf{ASR} $\downarrow$ & \textbf{OCR} $\uparrow$ & \textbf{Video QA} $\uparrow$ & \textbf{AVSR} $\downarrow$ & \textbf{AVQA} $\uparrow$ & \textbf{AVM} $\uparrow$ \\
    \midrule
    video-SALMONN   & 2.6\% & 37.8\% & 49.6\% & 7.7\% & 60.2\% & 79.7\% \\
    video-SALMONN without 5s-resolution & 2.5\% & 35.4\% & 47.2\% & 7.7\% & 57.2\% & 77.5\%\\
    video-SALMONN without 0.5s-resolution & 2.9\% & 37.1\% &  49.9\% & 8.3\% & 58.9\% & 80.6\% \\
    video-SALMONN without mixed training scheme & 2.6\% & 34.0\% & 46.9\% & 8.3\% & 54.0\% & 75.3\% \\
    video-SALMONN without diversity loss & 2.5\% & 36.8\% & 49.3\% & 7.7\% & 53.5\% & 78.6\% \\
    % video-SALMONN - causal encoder & 3.0\% & 34.9\% & 44.0\% & {8.0}\% & 37.1\% & 74.8\% \\
    video-SALMONN without MRC Q-Former & 3.3\% & 34.6\% & 42.7\% & 8.5\% & 45.3\% & 74.5\% \\
    % {video-SALMONN without synchronisation} & 40.6\% & 8.4\% & 53.4\% & 17.2\% & 50.5\% & 72.5\% \\
    {video-SALMONN without MRC Q-Former, sync. and div.}  & 3.1\% & 34.7\% & 36.0\% & 8.9\% & 44.6\% & 72.0\% \\
    \bottomrule
    \end{tabular}
    \label{tab:ablation1}
\end{table*}

\section{Results and Discussions}

\subsection{Main Results}

The results of video-SALMONN on the SAVE benchmark tasks are summarised in Table \ref{tab:singlemod} and Table \ref{tab:audiovisual} for single-modal and audio-visual tasks respectively. While other models can only perform a subset of SAVE tasks, video-SALMONN is the first single model that achieves competitive performance on all tasks with remarkably better performance on audio-visual tasks. In particular, video-SALMONN effectively achieves zero-shot audio-visual co-reasoning as an emergent ability, which is reflected by the performance on the two AVQA datasets, the AVSSD and AVM tasks.

On audio-based tasks in Table~\ref{tab:singlemod}, video-SALMONN obtains both the lowest WER and the highest SPIDEr scores compared to Whisper large-v2 and Video-LLaMA respectively. We do not report WER for Video-LLaMA as that is over 100\% due to a very high insertion rate.
% Note that the audio benchmark performance is close to the ones reported on the audio-only SALMONN \cite{salmonn} with much less training data.
% Further, with the aid of visual information, video-SALMONN achieves a lower WER on AVSR than Whisper-large-v2 in Table~\ref{tab:audiovisual}. 
On visual tasks, video-SALMONN demonstrates the best results on IC, OCR and Video QA, and on-par results on VQA with InstructBLIP fine-tuned on the same training set. In particular, the multi-resolution causal modelling in video-SALMONN yields over 25\% improvements compared to InstructBLIP even though the latter is fine-tuned on the same set of video data. This directly reflects the benefit of the MRC Q-Former.

On audio-visual tasks in Table \ref{tab:audiovisual}, video-SALMONN achieved 7.2\% relative WER reduction on the AVSR task compared to Whisper-large-v2. On the AVQA tasks, video-SALMONN achieved over 30\% accuracy improvements compared to the Video-LLaMA baseline which does not understand human speech, showcasing its comprehensive understanding ability for speech-audio-visual inputs. 

More importantly, video-SALMONN demonstrated a strong zero-shot audio-visual co-reasoning ability based on the AVM and AVSSD results compared to Video-LLaMA. 
% and is the only single-model system to our knowledge that can perform speech-image co-reasoning based on image-spoken QA (ISQA). 
Audio-visual co-reasoning (including speech-image co-reasoning) is an important yet challenging ability which requires the model to pay balanced attention to both audio and visual inputs as well as comprehending the intricate instruction beyond simply describing the inputs. This ability is especially enhanced in video-SALMONN by the unpaired audio-visual mixing strategy.
Such tasks were almost infeasible for any other audio-visual models so far, since they were unable to understand both speech and non-speech sounds, or were merely able to verbatim describe the input. Further discussion and qualitative analysis on audio-visual emergent abilities in addition to the audio-visual co-reasoning can be found in Section \ref{sec:emergent}.

\subsection{Ablation Studies}
\label{sec:ablation}

This section particularly focuses on the key structural novelty, including MRC Q-Former, the fine-grained synchronisation, as well as training techniques in video-SALMONN on selected SAVE benchmark tasks, as summarised in Table \ref{tab:ablation1}. 

First, we examine the effect of different resolution levels by training systems with either higher or lower resolutions. Modelling at different resolution levels results in a complementary outcome, where high resolution is better at ASR and AVSR and low resolution is better at OCR and Video-QA. The joint effect of the two resolutions gives the most balanced overall performance on all tasks.

Next, the effect of the mixed training scheme and diversity loss can be seen by comparing row 4 and row 5 to row 1 in Table \ref{tab:ablation1}. Both techniques provide improvements, particularly to audio-visual understanding tasks including AVQA and AVM, as the model pays balanced attention to both audio and visual streams as well as to different input frames.

Finally, we provide a comparison of the system without MRC Q-Former, and the system by further removing the temporal synchronisation, as shown in the last two rows of Table \ref{sec:ablation}. This is a fair comparison to highlight our novel model structure compared to Video-LLaMA under the same training dataset and the same frame rate. Without MRC Q-Former, while experiencing degradation across all tasks, the degradation in ASR, AVSR and Video-QA is the most obvious, as those tasks benefit the most from the multi-resolution design. By further removing the synchronisation, performances on AVSR and AVSSD degrade further due to the lack of cross-modal interactions at the feature level.

\begin{figure}[t]
    \centering
\includegraphics[scale=0.28]{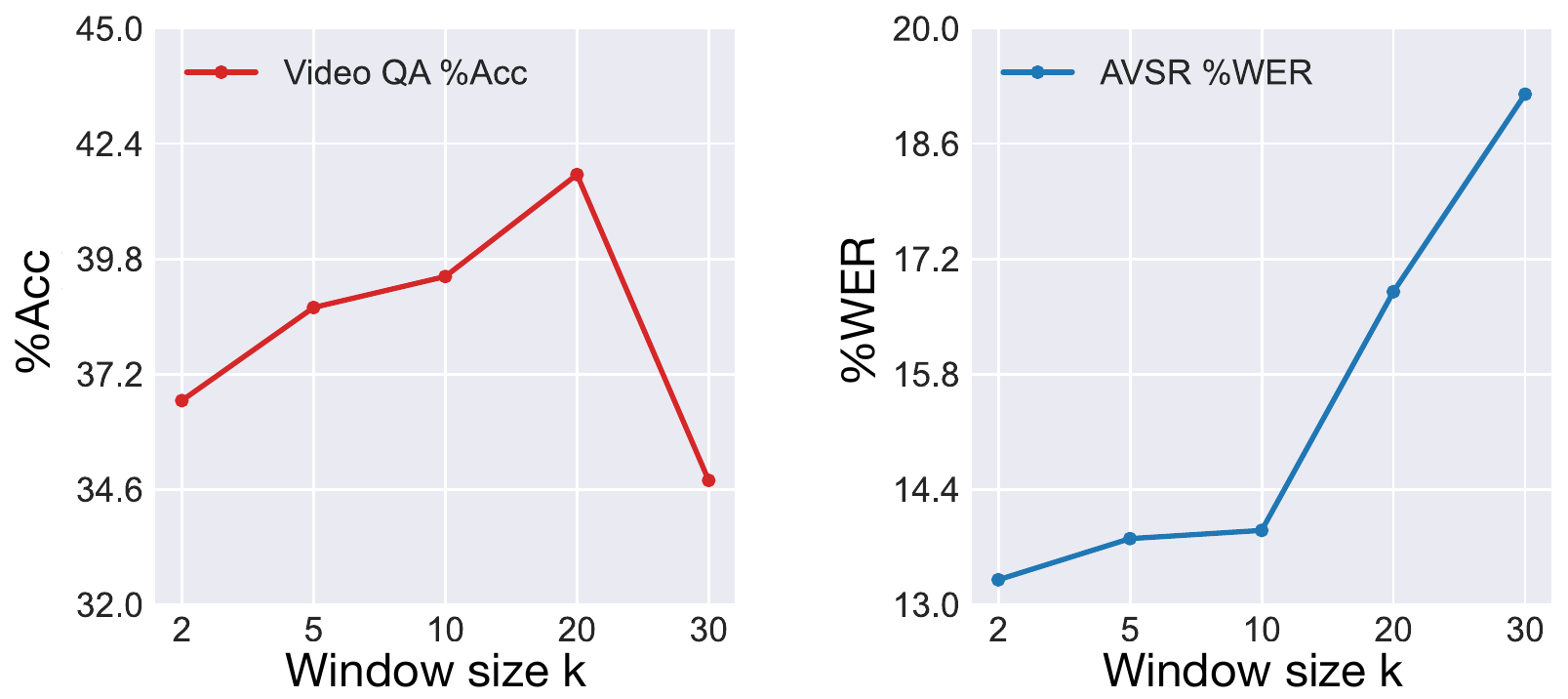}
    \vspace{-0.3cm}
    \caption{Influence of the window sizes $k$ to the model performance on video QA and AVSR. Results are from systems trained on 10\% randomly sampled data for efficient experiments.}
    \label{fig:resolution}
    \vspace{-0.3cm}
\end{figure}
\subsection{Analysis on Multi-resolution}
\label{sec:multires}
The MRC Q-Former extracts semantic information from the multimodal inputs at different time scales, which is necessary due to the nature of speech and visual inputs. This can be illustrated by plotting the influence on the performance of video-SALMONN on ASR and Video QA tasks against the number of frames $k$ in a window, as shown in Fig. \ref{fig:resolution}. For simplicity, only a single resolution level is used for these experiments. The ratio $N/k$ is kept constant which keeps the total number of output queries $C=W\times N$ unchanged for varying window sizes.

Speech contains temporally fine-grained information which requires high-resolution modelling to achieve better performance. Hence the WER decreases when the window size becomes smaller (\textit{i.e.} higher resolution). On the other hand, when the window size becomes smaller, fewer output tokens are used to encapsulate all the visual information within that window, causing performance degradation on video QA. Therefore, it presents a trade-off between speech and visual inputs about the granularity of the sliding windows, validating our motivation for the multi-resolution design.

% Figure \ref{fig:resolution} (c) clearly shows the importance of high temporal resolution in video modelling. The lowest FPS is equivalent to 8 frames per video, \textit{e.g.} Video-LLaMA, where over 24\% relative accuracy improvements are achieved using an FPS of 2. The best model trained on the full set is used with the same number of frames per window. While low accuracy is observed when the frame rate is low, increasing FPS beyond 1.0 only receives marginal improvements at the cost of having many more output queries sent to the LLM. 2.0 FPS was chosen as it made the audio and visual sequences have the most similar lengths, and hence easier for synchronisation.

To illustrate the functionality of each resolution level, we apply zero masks to the output query of one resolution level and observe the performance of another, as shown in Table \ref{tab:resolution2}. The system learns to split the functionality into two resolutions: the high resolution takes care of speech content-related information and the low resolution takes care of high-level information such as Video QA. This agrees with our findings from the ablation studies. Moreover, the complementarity of the two resolution levels is further processed by the LLM to achieve the best outcome.

\begin{table}[t]
    \centering
    \footnotesize
    \caption{Analysis of the effect of each resolution level reflected by ASR, IC and Video-QA tasks, with average cosine similarity between query vectors and word embeddings shown in brackets.}
    \vspace{0.2cm}
    \begin{tabular}{lccc}
    \toprule
    \textbf{Resolution level}    & \textbf{ASR} $\downarrow$ & \textbf{IC} $\uparrow$ & \textbf{Video QA} $\uparrow$ \\
    \midrule
    Both & 2.6\% & 89.6 & 49.6\% \\
    0.5s & 2.6\% & 35.8  & 14.4\% \\
    5.0s   & 100+\% & 23.0 & 41.9\% \\
    % 10s %15s & 100+\% & 79.3 & 46.1\% \\
    \bottomrule
    \end{tabular}
    \vspace{-0.3cm}
    \label{tab:resolution2}
\end{table}

\subsection{Analysis of the Diversity Loss}
Analysis of the effect of diversity loss is also performed using 10\% of the training data as shown in Figure \ref{fig:divloss}, and examples of cosine similarity matrices among output queries are shown in Appendix \ref{sec:div}. For ASR, the model is trained to include all the speech information in the audio sequence and the cosine similarity varies according to the length of the speech. For videos, the cosine similarity does not vary a lot for different video lengths, and hence diversity loss effectively acts as a way to encourage more diversified information to be captured. However, when a high $\lambda$ is employed, diverse output queries confuse the LLM and hence cause severe hallucination problems (\textit{e.g.} high insertion rate in WER) that degrades performance.
\begin{figure}[t]
    \centering
    \includegraphics[scale=0.24]{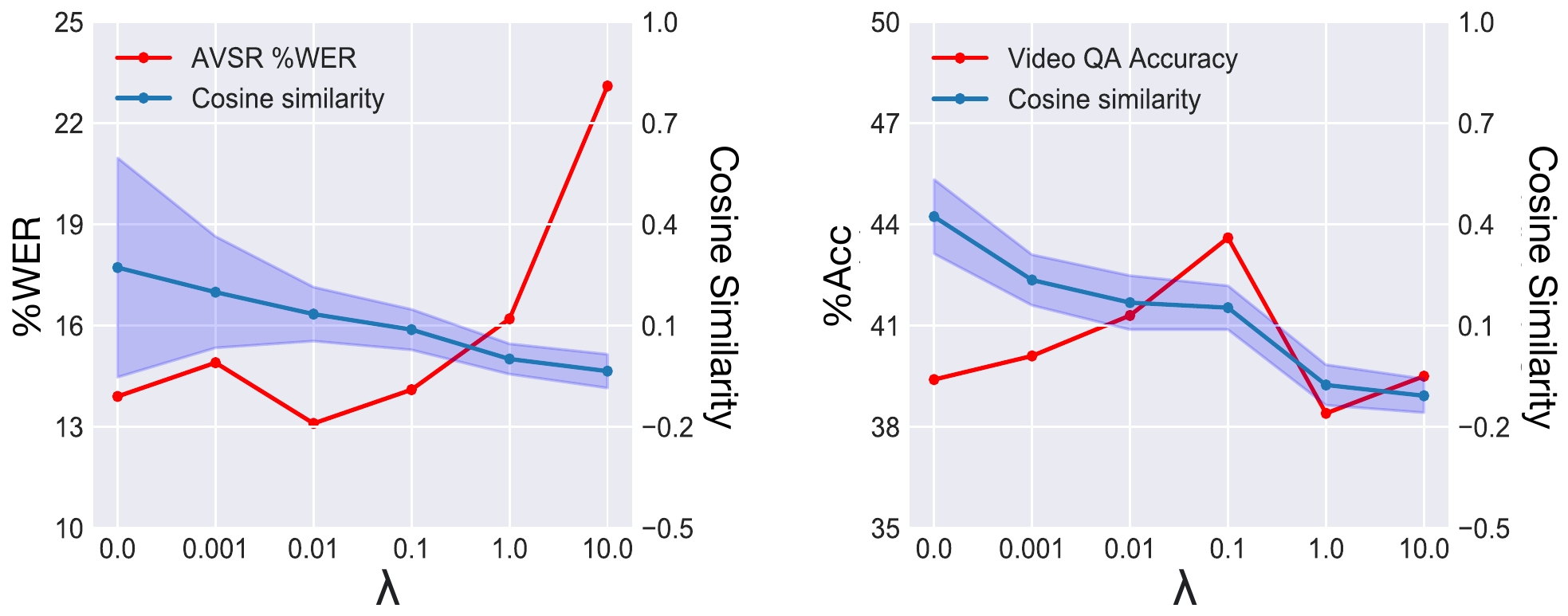}
    \vspace{-0.3cm}
    \caption{Variations of model performance by varying diversity loss factor, \textit{i.e.} $\lambda$ in Eqn. (\ref{eq:div}), on (a) AVSR (\%WER), and (b) Video QA (\%Accuracy). Variations of average cosine similarities among output query vectors are also shown under different $\lambda$'s.}
    \vspace{-0.3cm}
    \label{fig:divloss}
\end{figure}

\subsection{Emergent Speech-Audio-Visual Co-reasoning}
\label{sec:emergent}

% Speech is the most natural way of human-human interaction and is indispensable in understanding almost every video involving humans. 
% One of the key differences that set video-SALMONN apart from any existing av-LLM is the ability to accurately understand speech input. 
% In addition to the AVQA and AVM tasks that have already reflected the co-reasoning ability, many other unprecedented emergent abilities of video-SALMONN are shown via examples in Appendix \ref{casestudy}. 
%Similar to SALMONN \cite{salmonn}, 

In addition to objective measurements, we illustrate the unprecedented emergent speech-audio-visual co-reasoning abilities of video-SALMONN via examples in Appendix \ref{casestudy}. For instance,
video-SALMONN can answer questions in the speech about the image or video (see Fig. \ref{fig:eg1}). Benefiting from the mixed training scheme, video-SALMONN can write a coherent story based on unpaired audio and video (see Fig. \ref{fig:eg3}). More importantly,
in response to questions about why a movie clip is funny or romantic, video-SALMONN combines the video, dialogue between characters and background audio or music to generate a more encompassing and convincing answer (see Fig. \ref{fig:eg4} and \ref{fig:eg7}). Besides, video-SALMONN can understand the scene better by using knowledge from the speech, such as the species of a particular fish introduced in a documentary (see Fig. \ref{fig:eg5}). Moreover, the co-occurrence of speech and video events, such as attributing an utterance to a specific character (see Fig. \ref{fig:eg6} and Fig. \ref{fig:eg8}), can only be achieved by the dedicated structural design of video-SALMONN. 
% Finally, yet importantly, video-SALMONN with its multilingual abilities, can understand videos not only in English but also in Chinese, German and many other languages, significantly improving the language fairness for audio-visual LLMs.

\section{Conclusions}
%This paper proposes video-SALMONN as the first single unified model to achieve speech understanding in av-LLMs. Speech-driven structural design including the MRC Q-Former, the fine-grained synchronisation together with dedicated training strategies are proposed to enhance comprehensive video understanding abilities. Evaluated on the introduced AVE benchmark, video-SALMONN demonstrates superior performance compared to single-modal baselines, while achieving 25\% accuracy improvements on video-QA and over 30\% accuracy improvements on audio-visual QA compared to Video-LLaMA. Moreover, video-SALMONN showcases unprecedented audio-visual, and particularly strong speech-visual co-reasoning abilities, with remarkable cross-modal emergent abilities illustrated via examples.

This paper proposes video-SALMONN, the first single end-to-end av-LLMs that can understand all elements in video data, including visual frame sequence, speech, audio events, and music. To enhance the model's speech and comprehensive video understanding abilities, structural designs including MRC Q-Former, fine-grained synchronisation and a mixed training scheme are proposed. Evaluated on the introduced SAVE benchmark, video-SALMONN demonstrates superior performance compared to single-modal baselines, while achieving 25\% accuracy improvements on Video QA and over 30\% accuracy improvements on audio-visual QA compared to a strong baseline of Video-LLaMA. Moreover, video-SALMONN showcases unprecedented audio-visual, and particularly strong speech-visual co-reasoning abilities, with remarkable emergent abilities illustrated via examples.

\section*{Impact Statement}
Enabling speech understanding in av-LLMs marks an advancement towards achieving artificial general intelligence (AGI). By integrating speech input on top of existing non-speech audio and visual inputs, such a model would gain a holistic understanding of human interaction and the environment and is enabled to a broader range of applications. The potential positive impacts include:
%The approaches in this paper enable LLMs to understand speech, with the following potential positive impacts:
\begin{itemize}
    \item video-SALMONN enables more natural and intuitive interactions with technology, reducing the learning curve for users and making LLM-based technologies more approachable \textit{e.g.} for children and the elderly.
    \item video-SALMONN can potentially enhance the accessibility of LLM-based technologies, including those with motor impairments that make typing difficult.
    \item The video-QA demonstrates the potential of using video-SALMONN in academic presentations and educational applications to facilitate learning.
\end{itemize}
The approaches in this paper do not give rise to any additional potential biases beyond the ones directly inherited from the pre-trained model checkpoints used. The audio encoder and visual encoder might work worse for people from particular demographics. The framework also inherits biases from all the LLMs used in this paper. To mitigate potential biases, we clearly describe the nature of each dataset and provide clear and adequate references to all the resources we used for video-SALMONN.

The ability of video-SALMONN to understand speech in videos could lead to potential technology abuses like surveillance and eavesdropping. To counter this, we've consulted with legal experts to establish clear usage guidelines, reducing risks and addressing concerns, highlighting our dedication to responsible research sharing.

\bibliography{example_paper}
\bibliographystyle{icml2024}

%%%%%%%%%%%%%%%%%%%%%%%%%%%%%%%%%%%%%%%%%%%%%%%%%%%%%%%%%%%%%%%%%%%%%%%%%%%%%%%
%%%%%%%%%%%%%%%%%%%%%%%%%%%%%%%%%%%%%%%%%%%%%%%%%%%%%%%%%%%%%%%%%%%%%%%%%%%%%%%
% APPENDIX
%%%%%%%%%%%%%%%%%%%%%%%%%%%%%%%%%%%%%%%%%%%%%%%%%%%%%%%%%%%%%%%%%%%%%%%%%%%%%%%
%%%%%%%%%%%%%%%%%%%%%%%%%%%%%%%%%%%%%%%%%%%%%%%%%%%%%%%%%%%%%%%%%%%%%%%%%%%%%%%
\newpage
\appendix
\onecolumn
\section{Training Set and Benchmark Details}
\label{testset}

A range of datasets spanning audio and visual tasks are used in our experiments. Table \ref{tab:databenchmark} and \ref{tab:databenchmark2} summarise these datasets in detail, with individual descriptions and relevant prompt designs.

\begin{table}[H]
    \vspace{-0.3cm}
    \centering
    \footnotesize
    \caption{Dataset and benchmark details part 1}
    \begin{tabular}{lccp{3.8in}}
    \toprule
    \textbf{Data}     & \textbf{In Train} & \textbf{In SAVE} & \textbf{Description} \\
    \midrule
    LibriSpeech & Yes & Yes & LibriSpeech is an English audiobook data. The train-clean-100 and train-clean-360 splits were used for training, and test-clean was used in SAVE. Prompt example: ``Transcribe the speech into text." \\
    \midrule
    AudioCaps & Yes & Yes & AudioCaps is a widely used audio caption dataset containing 46k 10-second audio samples with manually annotated captions. Example prompt: ``Please describe the audio." \\
    \midrule
    LLAVA-150k & Yes & No & LLAVA-150k contain QA pairs generated using ChatGPT. Example prompt: ``What does the man hold in the image?"\\
    \midrule
    OCRVQA & Yes & No & OCRVQA is an OCR-based QA dataset containing questions mostly about printed words in an image. Example prompt: ``Who wrote this book?'' \\
    \midrule
    TextVQA & No & Yes & OCR-based QA dataset containing questions about various words in realistic scenes (\textit{c.f.} printed words). Example prompt: ``What is the brand of this camera?'' \\
    \midrule
    Flickr30k & No & Yes & Image caption dataset where each image is annotated with manual single-sentence descriptions. Example prompt: ``Describe this image in one short sentence.'' \\
    \midrule
    GQA & No & Yes & GQA consists of questions about various day-to-day real-world images. This involves reasoning skills about the objects in the image. Example prompt: ``What kind of device is on top of the desk?'' \\
    \midrule
    TextCaps & Yes & No & Image caption data particularly focusing on capturing text in the image. Only 80k samples were randomly selected for training. Example prompt: ``Describe the image.''\\
    \midrule
    MSVD-QA & Yes & No & MSVD-QA is a dataset with questions about real-world video clips. Example prompt: ``In the video, what is the man with long hair playing?'' \\
    \midrule
    NExT-QA & Yes & Yes & NExT-QA is a video QA dataset, particularly focusing on causal and temporal correlations. Example prompt: ``What does the girl in white do after bending down in the middle? Options/Choose one from: (Add choices here during inference)''. \\
    \midrule
    VideoChat & Yes & No & A GPT4-generated video QA dataset where the question mainly asks for detailed descriptions of the video. Example prompt: ``Provide a detailed description of the given video.'' \\
    \midrule
    AVSD & Yes & Yes & Audio-visual scene-aware dialogue data where questions are raised in turns about the video and the audio in the video. Example prompt: ``And then what happened?" and ``Is the man saying anything?'' \\
    \midrule
    Ego4D & Yes & Yes & An audio-visual dataset containing egocentric videos. Video descriptions were used as supervision signals which came from single-sentence short clip descriptions that were concatenated and refined using ChatGPT. Example prompt: ``Describe the video in detail.'' \\
    & & & 1000 video clips from the test set were used to make multiple choice questions by prompting ChatGPT with audio-visual caption and ASR transcription. \\
    \midrule
    How2 & Yes & Yes & An audio-visual speech recognition dataset containing videos explaining how to perform various tasks. Example prompt: ``Transcribe the speech into text, paying attention to both audio and video.''\\
    \bottomrule
    \end{tabular}
    
    \label{tab:databenchmark}
\end{table}

\begin{table}[h]
    % \vspace{-0.3cm}
    \centering
    \footnotesize
     \caption{Dataset and benchmark details part 2}
    \begin{tabular}{lccp{3.8in}}
    \toprule
    \textbf{Data}     & \textbf{In Train} & \textbf{In SAVE} & \textbf{Description} \\
    \midrule
    VGGSS & No & Yes & Sound source localisation data containing questions about the sound source in a 5-to-10-second video clip. Example prompt: ``What is the source of the sound?''\\
    \midrule
    Presentation-QA & No & Yes & A presentation video dataset labelled with slides text and speech transcriptions. 1000 video clips from the test set were used to make multiple-choice questions by prompting ChatGPT with slide content and ASR transcription. \\
    % \midrule
    % Bilibili-QA & No & Yes & A newly curated dataset for Chinese video understanding labelled with multiple choice questions by high-quality human annotators.\\
    \bottomrule
    \end{tabular}
   
    \label{tab:databenchmark2}
\end{table}

%%%%%%%%%%%%%%%%%%%%%%%%%%%%%%%%%%%%%%%%%%%%%%%%%%%%%%%%%%%%%%%%%%%%%%%%%%%%%%%
%%%%%%%%%%%%%%%%%%%%%%%%%%%%%%%%%%%%%%%%%%%%%%%%%%%%%%%%%%%%%%%%%%%%%%%%%%%%%%%

% \section{Comparison with other models}
% We extend our comparison of audio and visual single-modal abilities with two additional systems, namely the VAST \cite{vast} and the Valley \cite{valley} models. VAST is an av-LLM that can perform both audio caption and video understanding, whereas Valley is a pure visual LLM for video understanding. We compare the performance on 5 different tasks, including AAC on both AudioCaps and Clotho-v2 \cite{clotho}, IC on flickr30k and Video QA on MRSVTT \cite{msrvtt} and NExT-QA. The performances are summarised in Table \ref{tab:morecompare}.
% \begin{table}[h]
%     \centering
%     \caption{Comparison of video-SALMONN against VAST and Valley on audio and visual single-modal tasks. Note that for VAST, we only quote the numbers from the original paper \cite{vast} since we did not find the checkpoint for replication.}
%     \vspace{0.2cm}
%     \begin{tabular}{lccccc}
%     \toprule
%     System     & AAC (AudioCaps) $\uparrow$ & IC $\uparrow$ & MSRVTT-QA $\uparrow$ & NExT-QA $\uparrow$ \\
%     \midrule
%      VAST    &  76.9 & - & 46.8 & - \\
%      Valley & - & 30.8 & 45.7 & 24.6 \\
%      video-SALMONN & \textbf{81.7} & \textbf{89.6} & \textbf{51.5} & \textbf{49.6}\\
%     \bottomrule
%     \end{tabular}
%     \label{tab:morecompare}
% \end{table}

\newpage
\section{Examples of the AVQA Dataset}
\label{bilibiliqa}
\begin{figure}[H]
    \centering
    \includegraphics[scale=0.3]{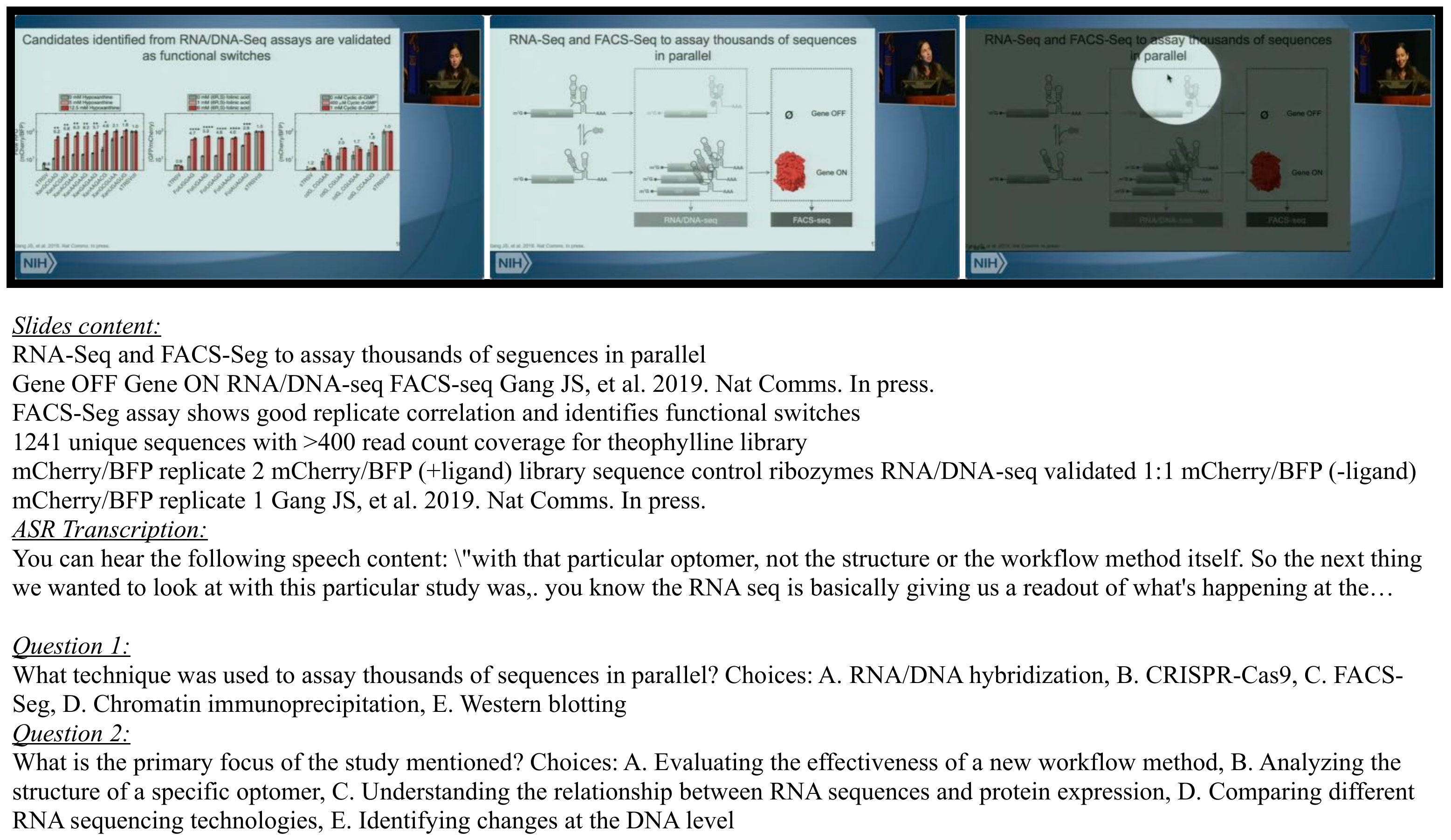}
    \caption{Example of Presentation-QA dataset.}
    \label{fig:presenQA}
\end{figure}

\begin{figure}[H]
    \centering
    \includegraphics[scale=0.3]{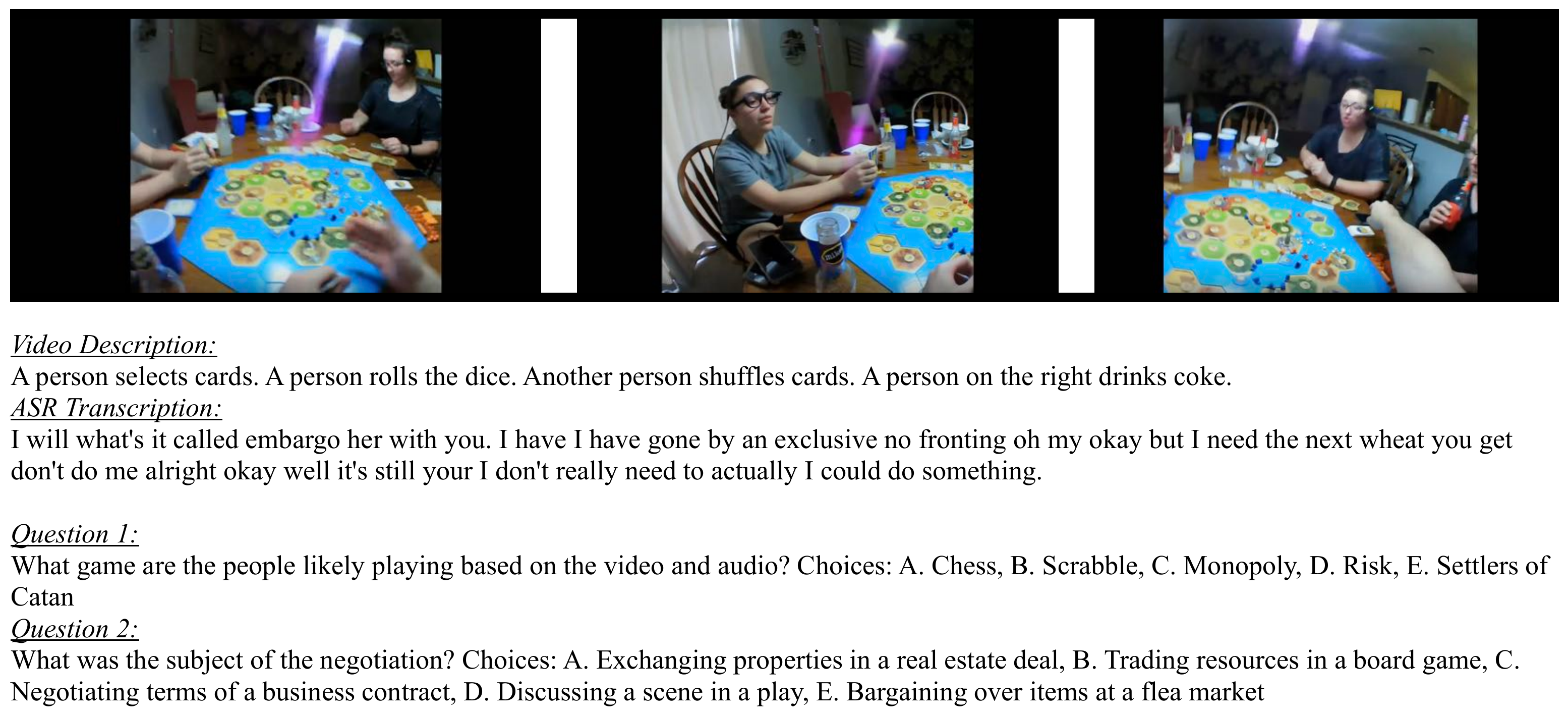}
    \caption{Example of Ego4D-QA dataset.}
    \label{fig:egoQA}
\end{figure}

The English AVQA datasets include Ego4D-QA and Presentation-QA with two examples shown in Fig. \ref{fig:presenQA} and \ref{fig:egoQA}.

% The bilibili-QA dataset was curated by manually labelling videos with multiple choice questions which reflect a complex reasoning about video contents. Three examples from this dataset are shown in Fig. \ref{fig:bilibili_eg1} and \ref{fig:bilibili_eg2}.
% \begin{figure}[h]
%     \centering
%     \includegraphics[scale=0.3]{bilibili_eg1.pdf}
%     \caption{Example of bilibili-QA dataset.}
%     \label{fig:bilibili_eg1}
% \end{figure}

% \begin{figure}[h]
%     \centering
%     \includegraphics[scale=0.3]{bilibili_eg2.pdf}
%     \caption{Example of bilibili-QA dataset.}
%     \label{fig:bilibili_eg2}
% \end{figure}

\section{Evaluation Details}
\label{evaluation}

{ASR and AAC are evaluated using word error rate (WER) and SPIDEr \citep{spider}, a combination of SPICE and CIDEr respectively. The evaluation of IC uses CIDEr following \citep{instructblip}. OCR, VQA, and Video QA are measured using top-1 accuracy. For OCR, the scoring follows \citep{textvqa} where each hit in the reference answer contributes 1/3 to the total hit. For VQA and Video QA, it is counted as correct if the reference answer exactly exists in the generated answer using word-by-word matching. It is needed to check the opposite answer doesn't exist for yes-or-no questions. In particular, during inference only, Video QA is formulated as an in-context multiple-choice task where the choices are given in the prompt, and one hit is counted only when the generated answer exactly matches the reference. The same measurement is taken for AVM. Furthermore, for AVSSD, as the reference answer is a full sentence, ChatGPT-assisted scoring is used to determine whether the generated answer is equivalent to the reference answer (see the prompt design in \ref{promptdesign}).}

\section{GPT Scoring Prompt Design}
\label{promptdesign}
As open-ended questions in VGGSS dataset contain full-sentence answers rather than one or two words, it is difficult to evaluate via string matching. Therefore, ChatGPT (GPT-3.5-turbo) was used to assist with the evaluation. Prompt designs for each task are described in Table \ref{tab:evalprompt}.
\begin{table}[h]
    \centering
    \caption{Prompt designs for ChatGPT-based evaluation. Note that \texttt{QUESTION} refers to the question, \texttt{HYPOTHESIS} is the model-generated answer and \texttt{REFERENCE} is the reference answer.}
    \begin{tabular}{lp{4.7in}}
    \toprule
     Task    &  Description\\
    \midrule
    VGGSS & Is the sound source mentioned in answer ``\texttt{REFERENCE}" the same as the sound source mentioned in answer ``\texttt{HYPOTHESIS}"? Answer ``Yes" if they are the same, and "No" if they are different or one does not mention the sound source.\\
    \bottomrule
    \end{tabular}
    % \caption{Prompt designs for ChatGPT-based evaluation. Note that \texttt{QUESTION} refers to the question, \texttt{HYPOTHESIS} is the model-generated answer and \texttt{REFERENCE} is the reference answer.}
    \label{tab:evalprompt}
\end{table}

% \section{EMD-based Cosine Similarity Metric}
% \label{emdexplain}
% To quantitatively analyse this multilingual ability, we apply a simplified version of the Earth Mover's Distance (EMD) based on cosine similarity as follows.
% \begin{equation}
%     S({U}_m, {V}_n) = \max_{F}\sum_i\sum_j f_{i,j}s(\mathbf{u}_i, \mathbf{v}_j)
% \end{equation}
% where $f_{i,j}$ are elements of $\mathbf{F}$ defining the flow matrix and ${U}_m, {V}_n$ are matrices containing the sequence of vectors dumped from a specific part of the model for each utterance, and $s(\mathbf{u}_i, \mathbf{v}_j)$ measures the cosine similarity between the two vectors. We enforce $f_{i,j}\in \{0, 1\}$, and the maximisation problem becomes a linear assignment problem which can be solved efficiently by the Hungarian algorithm. 

% Denoting $\mathcal{U}$ as the set of English utterances and $\mathcal{V}$ as the set of Chinese utterances where $V_i$ is the translation of $U_i$ into Chinese. The metric between utterances with the same meaning is
% \begin{equation}
%     \Bar{S}_\text{same} = \frac{1}{N} \sum_{n}^N S(U_n, V_n)
% \end{equation}
% and the metric between utterances with different meanings can be calculated as
% \begin{equation}
%     \Bar{S}_\text{diff} = \frac{1}{N^2}\sum_{m, n, m\neq n} S(U_m, V_n)
% \end{equation}
% We can replace $\mathcal{V}$ with $\mathcal{U}$ to calculate the average similarity between utterances with different meanings from the same language.

\section{Visualisation of Diversity Loss Effect}
\label{sec:div}
\begin{figure}[h]
    \centering
    \includegraphics[scale=0.25]{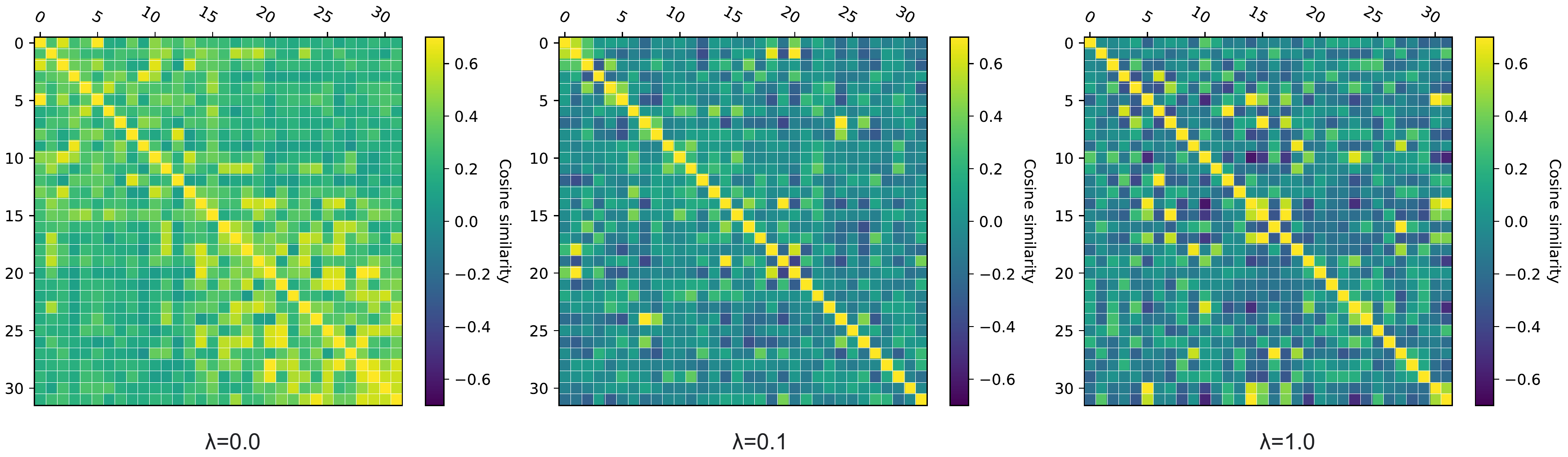}
    \vspace{-0.5cm}
    \caption{Visualisation of cosine similarity matrix with different diversity loss factors.}
    \label{fig:divvis}
\end{figure}

The cosine similarities among output query representations of the causal Q-Former under different diversity loss factors are shown in Fig. \ref{fig:divvis}.

\section{Additional Results on Lip Reading}
We further include the performance of video-SALMONN on the Oxford-BBC lip reading sentences 2 (LRS2) dataset. Results are shown in Table \ref{tab:lipreading}. video-SALMONN achieved better results than the Whisper baseline by a relative 7.5\% WER reduction.
\begin{table}[h]
    \centering
    \begin{tabular}{lcc}
    \toprule
    System & LRS2 \%WER & LSR2 + 0dB Gaussian noise \%WER\\
    \midrule
    Whisper large-v2 & 5.3 & 22.4 \\
    video-SALMONN audio alone & 5.1 & 22.4 \\
    video-SALMONN audio + video & \textbf{4.9} & \textbf{21.6} \\
    \bottomrule
    \end{tabular}
    \caption{\%WER on LRS2 lip-reading test set with clean speech, or with speech corrupted by 0dB Gaussian noise.}
    \label{tab:lipreading}
\end{table}

\section{Additional Results on MUSIC-AVQA}
We report our zero-shot MUSIC-AVQA results (without training on the MUSIC-AVQA dataset) in the following table, with a comparison to the AV-LLM \cite{audiovisual} and Video-LLaMA \cite{videollama}.

\begin{table}[h]
    \centering
    \begin{tabular}{lcc}
    \toprule
    System & MUSIC-AVQA Acc (\%) \\
    \midrule
    Video-LLaMA & 36.6\% \\
    AV-LLM & 45.2\% \\
    video-SALMONN & \textbf{52.6}\% \\
    \bottomrule
    \end{tabular}
    \caption{\%Acc on MUSIC-AVQA using Video-LLaMA, AV-LLM and video-SALMONN.}
    \label{tab:musicavqa}
\end{table}

\section{Comparison between Vicuna and Llama-2 as Backbone LLMs}
We provide the additional results using Llama-2 in contrast to Vicuna-v1.5 on SAVE in Table \ref{tab:llama2_1} and \ref{tab:llama2_2}.

\begin{table}[h]
    \centering
    \begin{tabular}{lcccccc}
    \toprule
    System & 	ASR	& AC & Video QA & IC & OCR & VQA \\
    \midrule
    video-SALMONN Vicuna-v1.5 & 2.6\% & 49.7\% & 49.6\% & 89.6\% & 37.8\% & 44.8\% \\
    video-SALMONN Llama-2 & 2.6\% & 50.6\% & 36.7\% & 91.6\% & 33.8\% & 45.4\% \\
    \bottomrule
    \end{tabular}
    \caption{Audio or visual-only tasks in SAVE for comparison between Llama-2 and Vicuna-v1.5 backbone LLM.}
    \label{tab:llama2_1}
\end{table}

\begin{table}[h]
    \centering
    \begin{tabular}{lcccccc}
    \toprule
    System & AVSR & AVQA (E) & AVQA (P) & AVSSD & AVM \\
    \midrule
    video-SALMONN Vicuna-v1.5 & 7.7\% & 49.8\% & 70.5\% & 47.6\% & 79.7\% \\
    video-SALMONN Llama-2 & 7.8\% & 40.6\% & 53.5\% & 48.6\% & 79.6\% \\
    \bottomrule
    \end{tabular}
    \caption{Audio-visual tasks in SAVE for comparison between Llama-2 and Vicuna-v1.5 backbone LLM.}
    \label{tab:llama2_2}
\end{table}

\section{Spotlight for Static Image}

We noticed that the performance of video-SALMONN on image tasks (e.g. VQA and OCR) may be limited by the lack of spatial resolution such that it is insufficient to extract details. To capture the finer details of an image, we make an extension to the MRC Q-Former by applying an image spotlight approach. We split the original image into a sequence of sub-images, and send the encodings of these sub-images to the MRC Q-Former in sequence. This is analogous to a video clip that scans the image patch by patch using a spotlight from the top left to the bottom right. The results of using the spotlight method (applied from the beginning of the instruction tuning) yielded better performance on OCR as shown in Table \ref{tab:spotlight}.
\begin{table}[H]
    \centering
    \begin{tabular}{lcccccc}
    \toprule
    System & ASR & AC & Video QA & IC & OCR & VQA \\
    \midrule
    InstructBLIP & - & - & 21.0\% & 84.5 & 36.5\% & 48.9\% \\
    video-SALMONN & 2.6\% & 49.7\% & \textbf{49.6}\% & \textbf{89.6} & 37.8\%	& 44.8\% \\
video-SALMONN + image spotlight & \textbf{2.6}\% & \textbf{50.6}\% & 49.1\% & 87.3 & 56.1\%	& 46.2\% \\
    \bottomrule
    \end{tabular}
    \caption{The SAVE benchmark single-modal task results using the spotlight of the static image.}
    \label{tab:spotlight}
\end{table}

Spotlight of the static image helped video-SALMONN to achieve much better results on OCR tasks, indicating that its performance on OCR is highly dependent on the image resolution. However, this slightly degrades the performance of video tasks. This can be due to the fact that the spotlight method has a different style of exploiting the input sequence from video frames, which slightly confuses the model.

\section{Case Studies}
\label{casestudy}
\begin{figure}[h]
    \centering
    \includegraphics[scale=0.3]{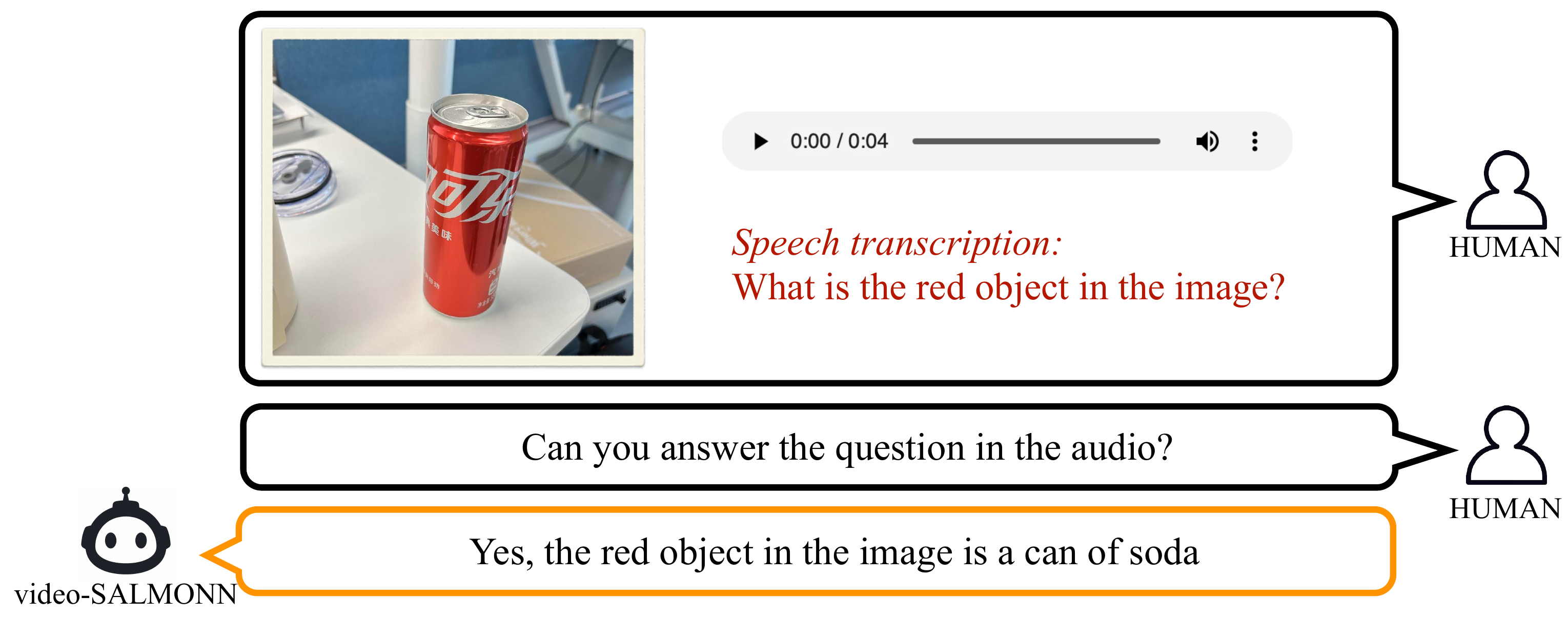}
    \caption{Case study 1 -- an example of the visual-spoken QA emergent ability.}
    \label{fig:eg1}
\end{figure}

\begin{figure}[h]
    \centering
    \includegraphics[scale=0.3]{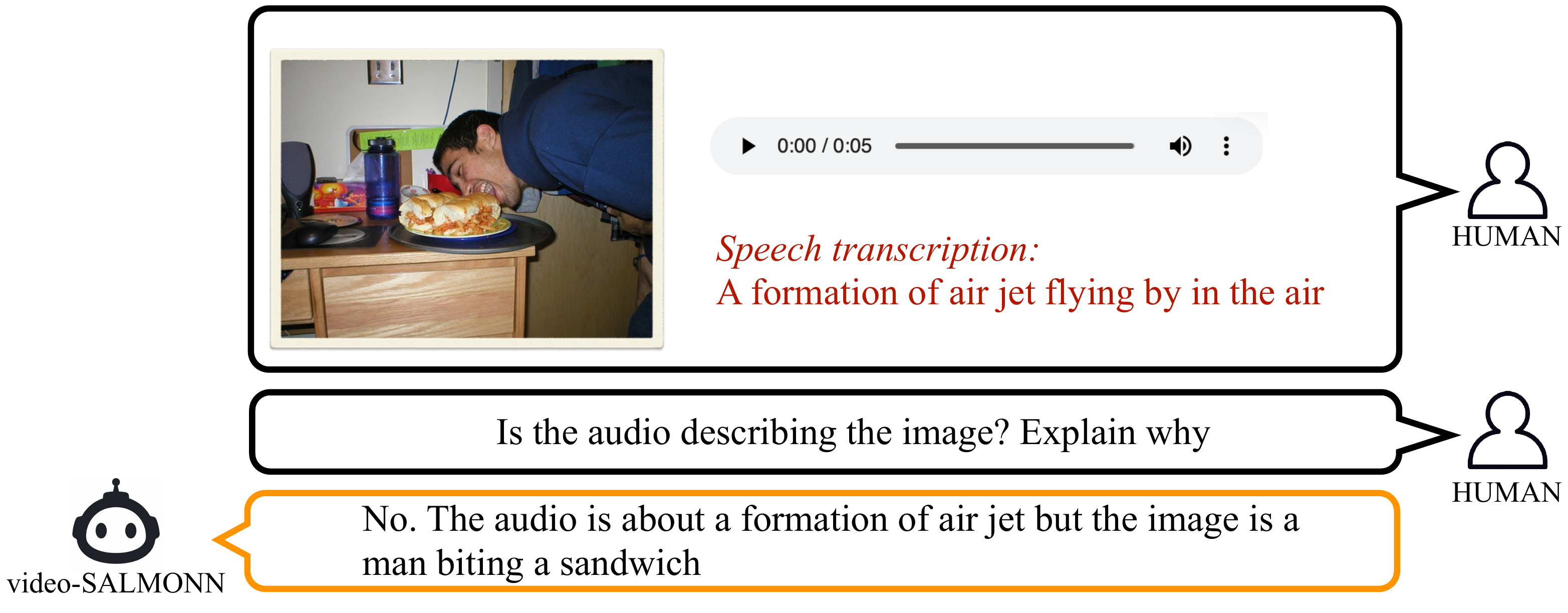}
    \caption{Case study 2 -- Audio-visual matching task with the request for explanation. During the benchmark test, the explanation was removed. The answer shows the understanding of both the speech and the image as well as the ability to perform reasoning based on them.}
    \label{fig:eg2}
\end{figure}

\begin{figure}[h]
    \centering
    \includegraphics[scale=0.3]{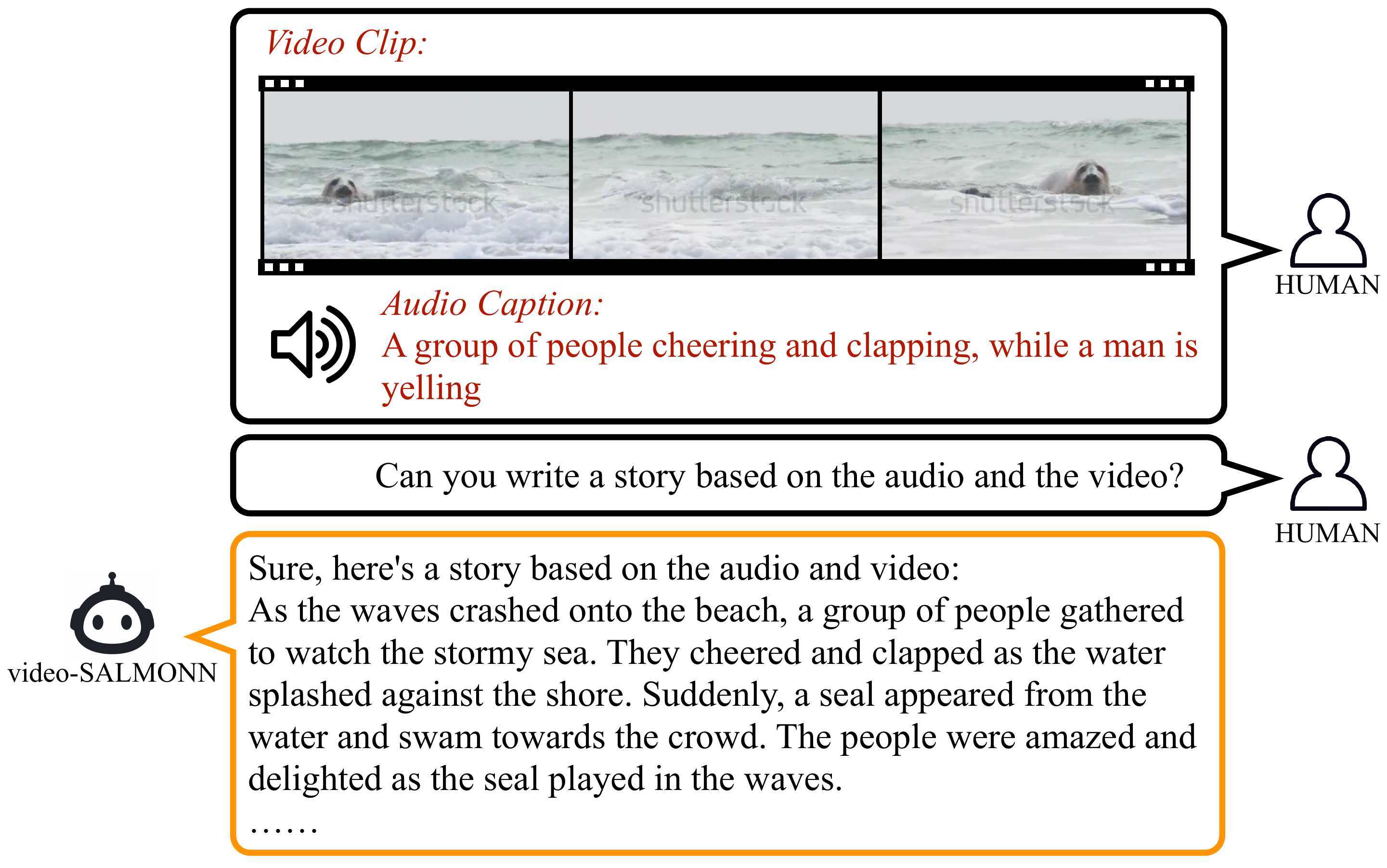}
    \caption{Case study 3 -- Storytelling task with a video clip and the audio came from a different source. The answer combines the audio event, such as cheering and clapping, coherently with the video content, such as the seal.}
    \label{fig:eg3}
\end{figure}

\begin{figure}[h]
    \centering
    \includegraphics[scale=0.3]{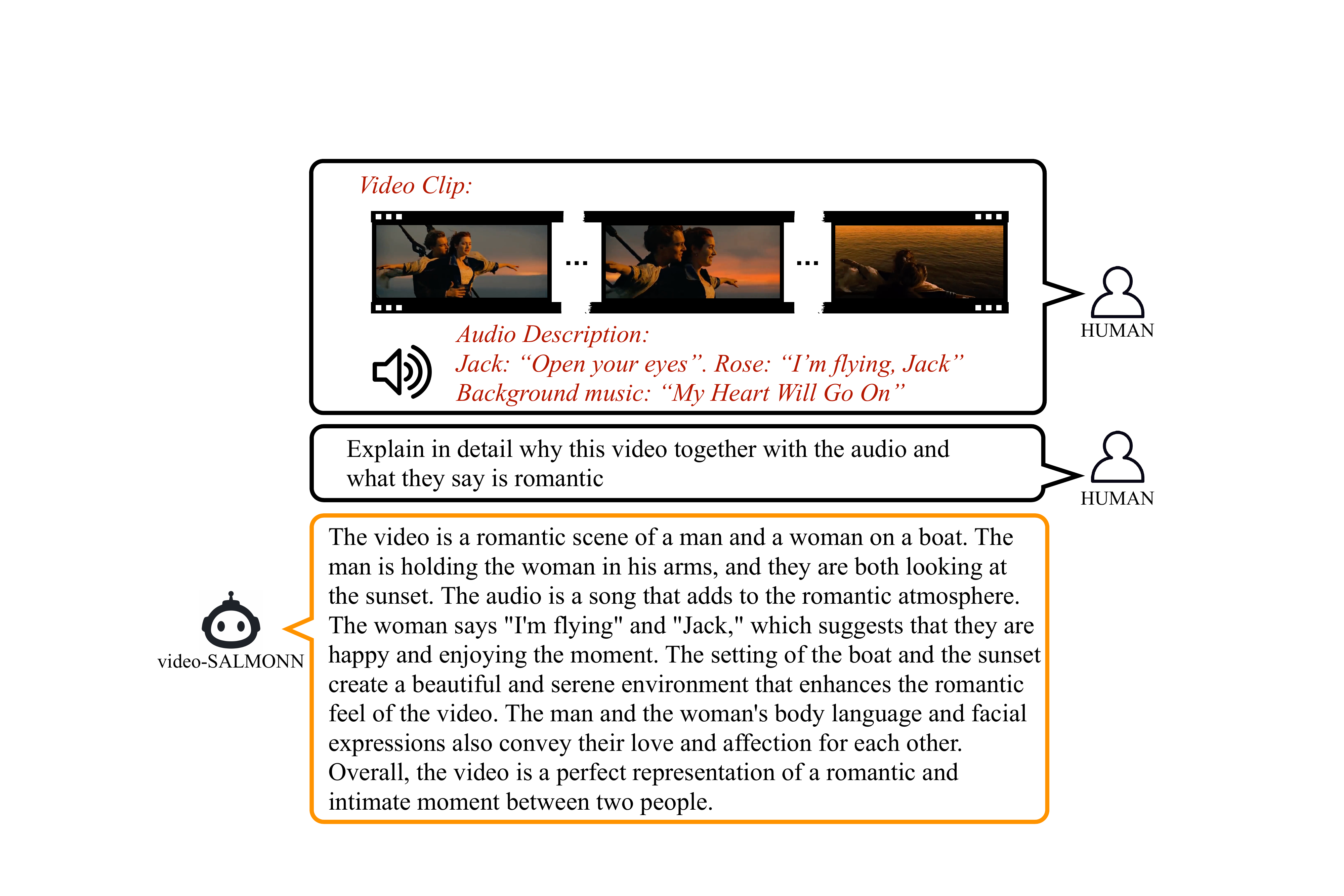}
    \caption{Case study 4 -- The famous scene in the movie \textit{Titanic} could be understood by video-SALMONN. The understanding combines the visual scene, the dialogue between characters, \textit{e.g.} ``I'm flying, Jack", as well as the background music to make the response comprehensive. It also reflects that the system knows the speaker by quoting the heroine's speech.}
    \label{fig:eg4}
\end{figure}

\begin{figure}[h]
    \centering
    \includegraphics[scale=0.3]{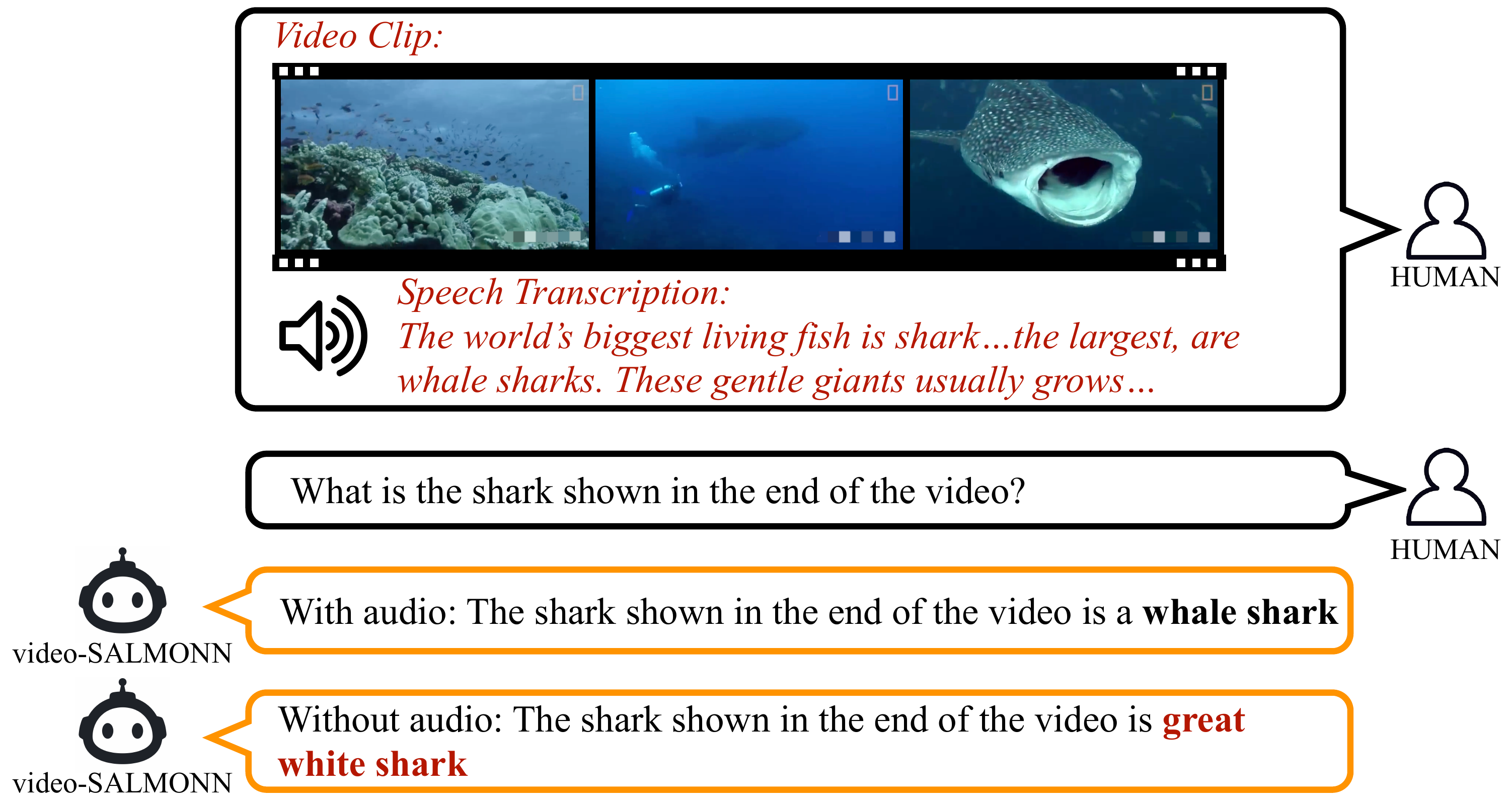}
    \caption{Case study 5 -- Demonstration of how speech content could provide knowledge for visual understanding. The system was clearly unable to identify the species of the shark without the help of the audio, and just made the most likely guess.}
    \label{fig:eg5}
\end{figure}

\begin{figure}[h]
    \centering
    \includegraphics[scale=0.3]{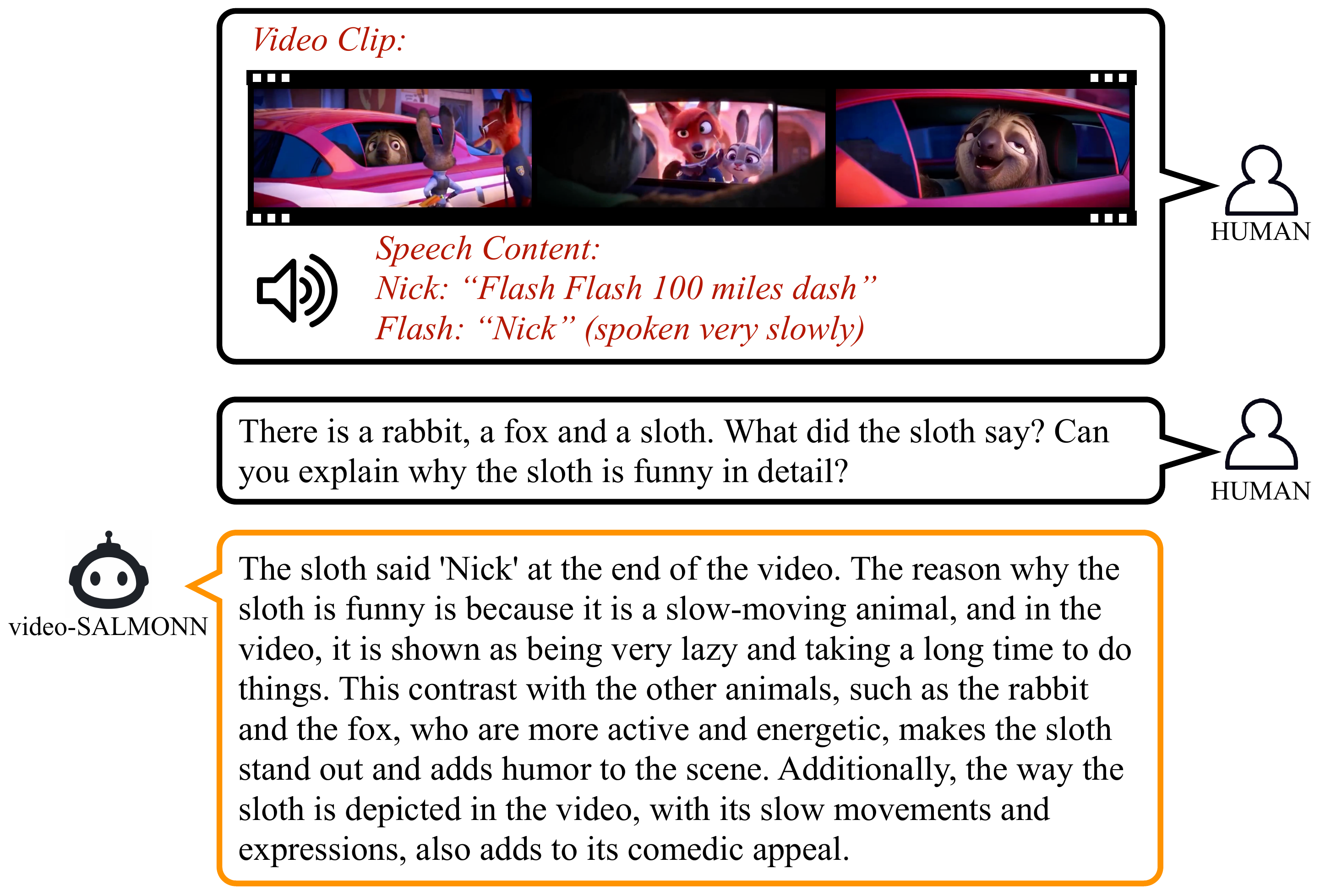}
    \caption{Case study 6 -- Demonstration of understanding cartoon clips about the amusing sloth character named ``Flash" in \textit{Zootopia}. video-SALMONN explained using both audio and video and accurately attributed the word ``Nick" to the sloth.}
    \label{fig:eg6}
\end{figure}

Six cases are illustrated in Fig. \ref{fig:eg1} to Fig. \ref{fig:eg6}.

\begin{figure}[h]
    \centering
    \includegraphics[scale=0.3]{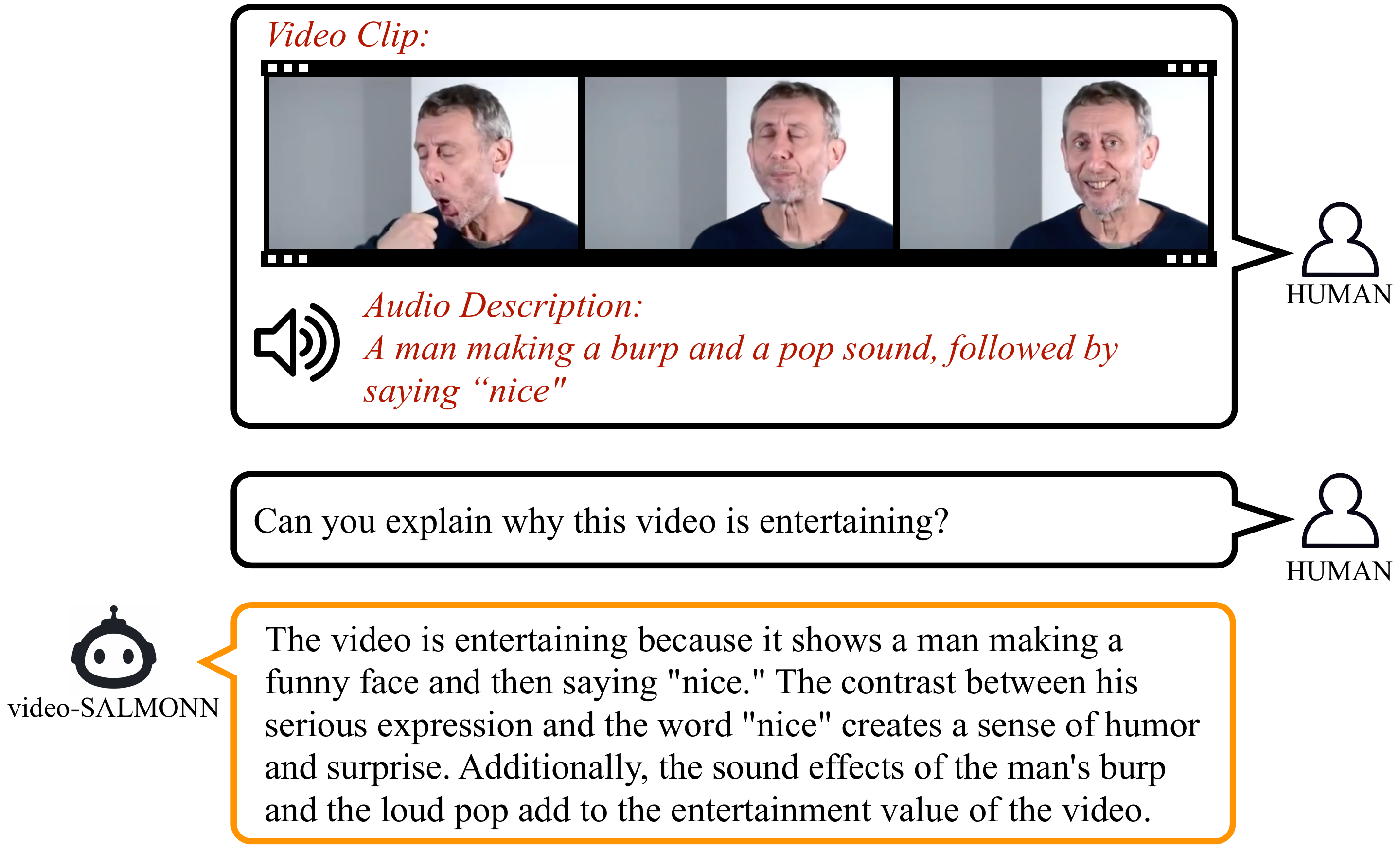}
    \caption{Case study 7 -- Demonstration of video-SALMONN using audio, speech and video to explain why a specific meme is interesting. The explanation includes the funny sound, the word being said with the facial expression. }
    \label{fig:eg7}
\end{figure}

\begin{figure}[h]
    \centering
    \includegraphics[scale=0.3]{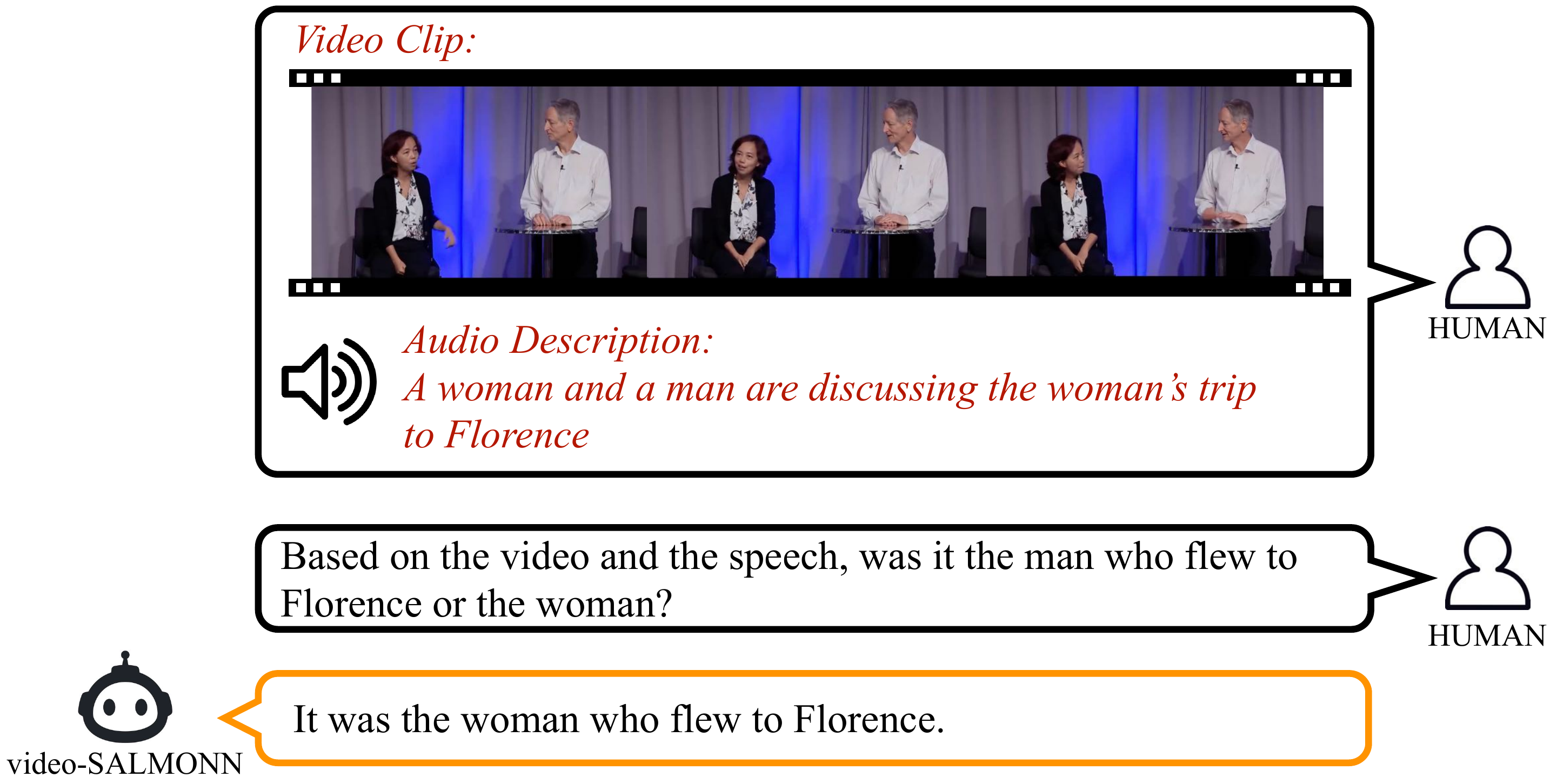}
    \caption{Case study 8 -- Demonstration of video-SALMONN using audio, speech and video to understand the speech content about who flew to Florence. Without the video content, it is difficult to infer who we are referring to.}
    \label{fig:eg8}
\end{figure}

\end{document}